\documentclass{article}
\usepackage{ijcai23}

\usepackage{times}
\usepackage{soul}
\usepackage{url}
\usepackage[hidelinks]{hyperref}
\usepackage[utf8]{inputenc}
\usepackage[small]{caption}
\usepackage{graphicx}
\usepackage{amsmath}
\usepackage{amsthm}
\usepackage{booktabs}
\usepackage{algorithm}
\usepackage{algorithmic}
\usepackage[switch]{lineno}

\title{Story Shaping: Teaching Agents Human-like Behavior with Stories}

\author{
Xiangyu Peng\footnote{Equal contributions} 
\and
Christopher Cui\footnotemark[1]
\and
Wei Zhou
\and
Renee Jia
\and
Mark Riedl 
\affiliations
Georgia Institute of Technology
\emails
\{xpeng62,  ccui46, wzhou322, rjia35\}@gatech.edu,
riedl@cc.gatech.edu
}

\usepackage{latexsym}
\usepackage{multirow}
\usepackage{booktabs}
\usepackage[normalem]{ulem}

\usepackage[utf8]{inputenc}
\usepackage{amsmath}
\usepackage{paralist, tabularx}

\usepackage{tikz}
\usepackage{enumitem}

\definecolor{brightcerulean}{rgb}{0.11, 0.67, 0.84}

\newcommand{\sysnameq}{Q*BERT-S}

\usepackage{graphicx,calc}
\newlength\myheight
\newlength\mydepth
\settototalheight\myheight{Xygp}
\usepackage{amssymb}

\newcommand{\baselinerobotsymbol}[0]{\text{\smash{\raisebox{-3pt}{\includegraphics[height=10pt]{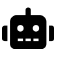}}}}}
\newcommand{\humanrobotsymbol}[0]{\text{\smash{\raisebox{-3pt}{\includegraphics[height=10pt]{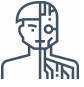}}}}}

\begin{document}

\maketitle

\begin{abstract}
Reward design for reinforcement learning agents can be difficult in situations where one not only wants the agent to achieve some effect in the world but where one also cares about {\em how} that effect is achieved.
For example, we might wish for an agent to adhere to a tacit understanding of commonsense, align itself to a preference for how to behave for purposes of safety, or taking on a particular role in an interactive game.
Storytelling is a mode for communicating tacit procedural knowledge.
We introduce a technique, {\em Story Shaping}, in which a reinforcement learning agent infers tacit knowledge from an exemplar story of how to accomplish a task and intrinsically rewards itself for performing actions that make its current environment adhere to that of the inferred story world.
Specifically, Story Shaping infers a knowledge graph representation of the world state from observations, and also infers a knowledge graph from the exemplar story.
An intrinsic reward is generated based on the similarity between the agent's inferred world state graph and the inferred story world graph.
We conducted experiments in text-based games requiring commonsense reasoning and shaping the behaviors of agents as virtual game characters.
\end{abstract}

\begin{figure}[tbh!]
    \centering
    \includegraphics{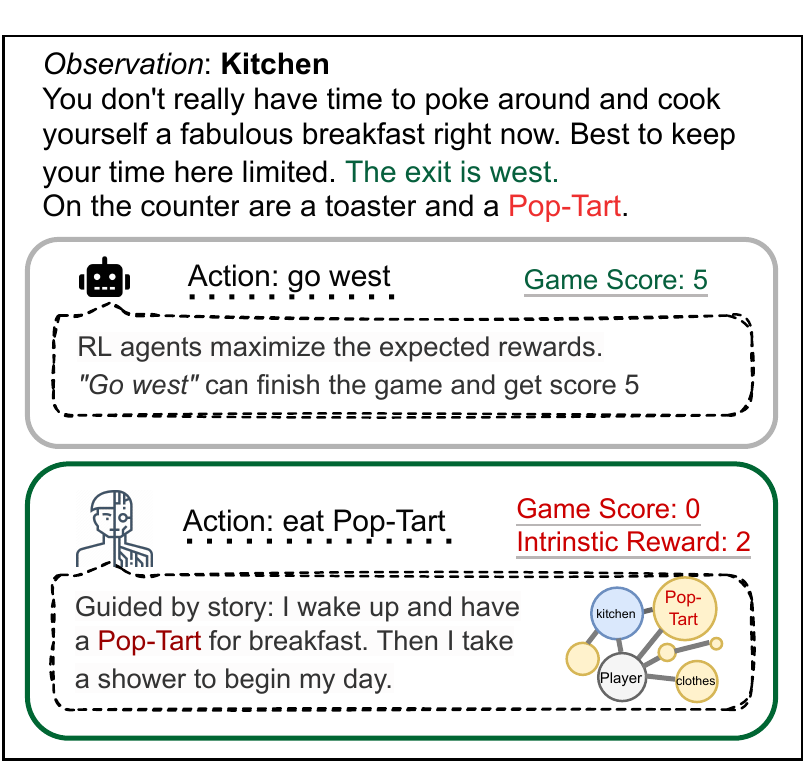}
    \caption{Excerpt from the text game, 
    \texttt{9:05}, with actions from
    vanilla reinforcement learning agent represented by \baselinerobotsymbol, 
    and RL agent with \textbf{Story Shaping} by \humanrobotsymbol.
    \baselinerobotsymbol \hspace{0.05cm} takes action ``\textit{go west}'', which only maximizes the expected rewards and ignores the commonsense knowledge in the environment.
    \humanrobotsymbol \hspace{0.05cm} instead chooses to do something more similar to what a human would do---eat a Pop-Tart---based on a story written in natural language, which may be provided or created automatically by our system.}
    
    \label{fig:fig1}
\end{figure}

\section{Introduction}

Reinforcement Learning (RL) is a class of techniques whereby an agent learns how to carry out a sequential task through repeated interaction with the environment. 
The agent is given a reward for achieving certain states in the environment, executing certain actions, or causing the world to transition between particular pairs of states.
RL is thus a process of learning to maximize expected future rewards. 
The design of the reward signal---when the reward (or penalty) is given and how much---determines what the optimal behavior for the agent should be.

Reward design can be especially difficult in situations where one not only wants the agent to achieve some effect in the world but where we also care about {\em how} that effect is achieved.
For example, we may want the agent to carry out the task in a way that adheres to social norms during execution.
This can make it safer for humans to work with AI-driven systems because they conform to our expectations and preferences~\cite{Frazier2020LearningNF,NahianTraining2021,ammanabrolu2022aligning}.
We may wish for our agents to demonstrate commonsense knowledge during execution~\cite{dambekodi20commonsense}.
In computer game environments, we may wish to have AI-driven agents 
role-play different types of characters with different interests or ways of accomplishing things~\cite{urbanek2019learning,ammanabrolu2021motivate}.

We introduce a new technique, \textbf{Story Shaping}, for specifying preferences over an agent's behavior.
In Story Shaping, a reward designer specifies {\em how} to perform a task by providing an example story.
Stories are efficient means for transferring tacit procedural knowledge between people.
Storytelling is a mode of communication wherein details are abstracted away under the assumption that the recipient shares a common base of knowledge with which to reconstruct details as necessary for comprehension~\cite{sternberg02}.

Story Shaping is a process whereby the agent reverse-engineers implicit state information from the exemplar story and constructs a rich, {\em intrinsic} reward signal that guides it toward behavior that makes its environment more closely resemble the implicit world of the story. 
Specifically, 
the agent infers a {\em knowledge graph}  from the story, consisting of $\langle subject, relation, object\rangle$ triples for every relation that can be inferred from the story.
The agent also extracts relations from its observations to construct a knowledge graph summarizing the operating environment.
The agent rewards itself for how similar its current world state representation is from the desired representation derived from the exemplar story.

Story shaping is related to {\em Learning from Demonstrations} (LfD). 
In LfD, the agent is provided with a set of traces, typically enacted by humans in the same environment.
The learning agent must reconstruct the latent policy that human demonstrators were following. 
Demonstrations are typically assumed to be complete (no missing steps) and performed in the same---or very similar---environment that the agent also operates in.
In contrast, Story Shaping 
does not assume the stories are complete nor executable, and does not assume they reference the same environment.
Instead, stories may reference the essential objects, locations, or activities but leave out details.
Additionally, because stories do not reference the exact environment the agent inhabits, objects and locations may differ or be missing, and actions may not be executable.

We experiment with Story Shaping in text games where observations and actions are presented entirely in text. 
Text games are partially observable environments that have large state and action spaces and often involve puzzles requiring long-range causal dependencies~\cite{hausknecht19}. 
Text games have also been demonstrated to transfer to visual and real-world domains~\cite{Wang2022Science,ALFWorld20}.
We conduct experiments across three game platforms that either (a)~require commonsense, or (b)~showed that our agent is able to role-play in a more human-like manner.
Additionally, we show that story-shaped agents can adapt their behavior to different character preferences.

\section{Background and Related Work}

Text games are turn-based games where the player 
must read human-written natural language (typically English) descriptions of the local environment and respond with short textual action descriptions.
A text game can be defined as a partially-observable Markov Decision Process: $\langle S, P, A, O, \Omega,R,\gamma\rangle$, representing the set of environment states, conditional transition probabilities between states, the vocabulary or words used to compose text commands, observations, observation conditional probabilities, reward function, and discount factor, respectively. 
The transition and observation probabilities, $P$ and $\Omega$, are typically unknown to the agent.
Observations are text sequence, and 
actions are composed of one to five tokens from a vocabulary.
The RL agent attempts to learn a policy $\pi(o)$ $\rightarrow a$ that maximize future expected reward.

\paragraph{Knowledge Graphs for Text Games.}
A knowledge graph is a set of $\langle subject, relation, object\rangle$ tuples. 
Knowledge-graph based reinforcement learning agents have been shown to be state-of-the-art in text-based games~\cite{ammanabrolu2018playing,ammanabrolu2020bringing,ammanabrolu2020graph,ammanabrolu2020avoid,xu2020deep,peng2021inherently}.
These agents infer objects and relations from text observations and use this knowledge graph as a long-term memory of the world state as a means of handling partial observability. 
We build off the KG-A2C~\cite{ammanabrolu2020graph} agent architecture, which uses the ALBERT~\cite{lan2019albert}
language model to infer objects and relations from the text observation, and a graph attention network to generate action sequences.
Whereas KG-A2C uses the knowledge graph to represent world state and filter actions, our Story Shaping approach also uses the knowledge graph to compute a dense reward signal, showing that KG-based reinforcement learning has additional untapped potential.

\begin{figure*}[tbh]
    \centering
    \includegraphics[width=\textwidth]{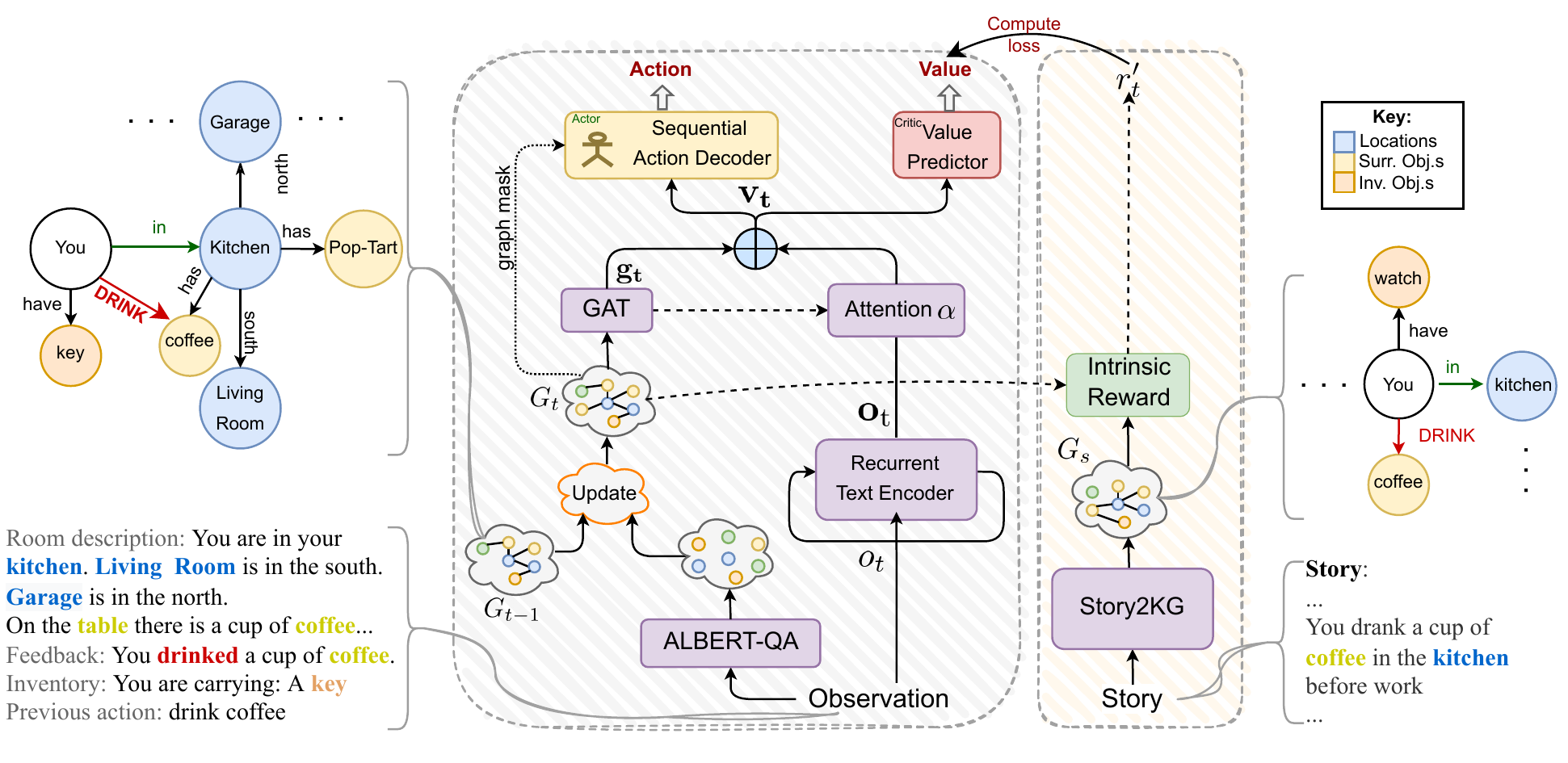}
    \caption{Knowledge graph extraction and the architecture of RL agent with Story Shaping at step $t$.
    }
    \label{fig:pipeline}
\end{figure*}

\paragraph{Intrinsic rewards.} 
Intrinsic rewards 
provide qualitative guidance~\cite{schmidhuber1991possibility,oudeyer2007intrinsic,barto2013intrinsic} for exploration and push an agent to get a specific behavior without any direct feedback from the environment.
These rewards can take many forms, such as a comparison between the agent’s predictions and reality~\cite{stadie2015incentivizing,burda2018large,kim2019curiosity}, or the performance on self-generated goals~\cite{vezhnevets2017feudal,levy2018hierarchical,nachum2018data,nair2018visual,pong2019skew}.
Ammanabrolu {\em et al.}~\shortcite{ammanabrolu2020avoid} intrinsically rewards a text-game playing agent for adding nodes and edges to a knowledge graph.
Our technique intrinsically motivates the agent to explore states that add nodes and edges that are anticipated by the exemplar story.
Related, the {\em learning from stories} technique by Harrison {\em et al.}~\shortcite{harrison2016towards} uses stories to guide RL agents. However, it requires dozens of exemplar stories and each event is treated as a goal in a modular hierarchical policy; we only require a single story and generate a single unified policy.

\begin{table}[t]
\centering
\footnotesize
\setlength\tabcolsep{0.1pt} %
\begin{tabular}{p{0.99\linewidth}}
\toprule
\textbf{Game: \texttt{9:05}}  \\
\textbf{First observation:}\\
 I wake up in the morning. A bathroom lies to the south, while a door to the east leads to the living room. On the end table are a telephone, a wallet and some keys. The phone rings.\\
\textbf{Goal: GO TO WORK}. \\
\midrule
\midrule
\textbf{Human-written story about a routine:}  \\
Upon waking in the morning, I start my day with a \underline{\textit{Pop-Tart}} breakfast, followed by a \underline{\textit{shower}} before commencing work.\\
\midrule

\textbf{Human-written story about a Persona (\texttt{refined man}):}\\
Upon waking, I first tend to my personal hygiene by taking a \underline{\textit{shower}} and using the \underline{\textit{toilet}}. 
After, I change into appropriate\underline{ \textit{clothes}} before having \underline{\textit{breakfast}}. 
I then leave my home to begin my workday.\\

\midrule

\textbf{Automatically generated story by ChatGPT:}\\
I will likely take a \underline{\textit{shower}} in the bathroom to the south, get \underline{\textit{dressed}}, and check my wallet and \underline{\textit{keys}} to make sure I have everything I need for the day. I may also \uwave{\textit{take a cup of coffee}} before leaving my home to go to work.\\

\bottomrule

\end{tabular}
\caption{Exemplar stories constructed by human or ChatGPT. 
Words \underline{underlined} indicate actions and entities that can be taken in the game.
Words with a wavy \uwave{underline} indicate actions or entities that are invalid or not allowed in the game.
}
\label{tab:ex_story}
\end{table}

\section{Story Shaping}
\label{sec:methods}

{\em Story Shaping} facilitates a RL agent's ability to learn implicit knowledge from an exemplar story about a task and rewards itself for actions that bring the operating environment more in alignment with the inferred story world.
Our technique starts with a given  exemplar story (which can also be automatically generated), which the agent
transforms into a knowledge graph, referred
as \textit{Story KG} (\S\ref{sec:story}). 
During RL training, as the agent explores the game world, it builds an internal state knowledge graph, called the \textit{World KG} 
(\S\ref{sec:game_kg}). 
Then the agent is updated using intrinsic rewards, calculated based on the similarity between the World KG and the Story KG 
(\S\ref{sec:intrinsic}).
The technique is overviewed in Figure~\ref{fig:pipeline}.

\subsection{The Exemplar Story}
\label{sec:story}

A natural language story is provided as an exemplar of the behavior the agent is to enact.
Table~\ref{tab:ex_story} shows some possible story exemplars for ``going to work in the morning''.
We assume that a person is providing the story as a high-level description of what the agent should attempt to achieve or to describe how a task should be achieved. 
The story can also be generated by an automated story generation system \cite{peng2021guiding,goldfarb2020content}
or by prompting a large language model such as ChatGPT\cite{chatgpt} to describe a typical way of doing something
(More details about prompt engineering for ChatGPT can be found in Appendix~\ref{app:chatgpt}).

Similar to the last example in Table~\ref{tab:ex_story}, 
stories can leave details about the environment out.
The person or entity that provides the story, whether human or large language model, may be unaware of the exact parameters of the operating environment of the agent. 
For example, ``coffee'' is not an entity in the game ``9:05'', but can be present in the exemplar example.
This is an intentional benefit of using story exemplars; stories can be thought of as compact descriptions that focus on salient details with the assumption that the recipient shares common knowledge with the storyteller and can thus reconstruct the more fine-grained
details.

The purpose of the exemplar story, however, is not to be a demonstration, but to generate 
an intrinsic reward signal that guides the reinforcement learning agent to act in ways that align the operating environment with the implicit world depicted in the story, 
thereby ``shaping'' the agent's behavior.
Before the agent begins training,
the exemplar story is converted into a knowledge graph---called the \textit{Story KG}---taking the form of RDF triples~\cite{klyne2004resource} of $\langle subject,relation,object\rangle$.
Story KG is an explicit and persistent memory of entities mentioned from the story.
This knowledge graph contains the entities and relations directly extracted from the exemplar story's text, as well as additional world details that can be inferred from the events in the story.

To acquire the Story KG, we train a Semantic Role Labeling (SRL)~\cite{gildea2002automatic}
model on VerbAtlas~\cite{di2019verbatlas} to provide the automatic identification and labeling of argument structures of the story.
VerbAtlas is a linguistic resource that provides semantic annotations on sentences based on the verb and how it is used. 
Verbs are important parts of stories because they convey action that change the state of the story world.
For example,``\textit{I drink coffee in the kitchen}'' is firstly processed by VerbAtlas SRL to obtain ``\texttt{\{'verbatlas': 'DRINK', 'description': '[ARG0: You] [drink.01: drink] [ARG1: coffee] [ARGM-LOC: in the kitchen]\}}''.
Then the themes and attributes are used as entities and VerbAtlas frames are used as edges, such as  $\langle$ \texttt{You, DRINK, coffee}$\rangle$.
We also incorporate location and time data into the knowledge graph (i.e., $\langle$ \texttt{You, in, kitchen}$\rangle$).
Training details can be found in Appendix~\ref{app:verbatlas-srl}.
Incorporating the VerbAtlas frame name of the verb as a relation satisfies three needs. 
First, it acts as a placeholder for the commonsense effects when they are not otherwise known.
For example, the coffee is in the state of having been drunk. 
Second, it acts as a record of actions that have been taken in the world---the world is one in which those actions occurred.
Third, some actions don't have {\em positive} effects that create relations, e.g., destroying something removes objects and relations because the object ceases to exist.

Story KG captures the positive relations at the end of the exemplar story, including the record of events as a placeholder for implicit effects.
An important attribute of this approach is that it intentionally does not capture the order of events.
The agent will learn through trial-and-error whether there are ordering constraints.
There may be events in the story that can be carried out in different orders or must be carried out in different orders depending on environmental conditions. 
Events may be missing completely (e.g., the story doesn't explicitly say to move from the bedroom to the kitchen).
Furthermore, some events may be impossible (e.g., there is no coffee in the kitchen). 
Instead, the story KG provides the key elements that the agent should encounter, and the agent receives more rewards for encountering and doing as many of them as possible.

\subsection{The Reinforcement Learning Agent}
\label{sec:rl_agent}

We consider the standard reinforcement learning setting
where an agent interacts with a text game environment over a number of discrete time steps.
State-of-the-art approaches to RL in text environments use a knowledge graph as an external, persistent memory of the world state~\cite{ammanabrolu2018playing,ammanabrolu2020graph,ammanabrolu2020avoid}.
As the agent explores the game world, a knowledge graph---called the {\em World KG}---is constructed and used as state representation.
The RL agent is trained via the Advantage Actor Critic (A2C) \cite{mnih2016asynchronous} to maintain a policy $\pi\left(a_t \mid s_t ; \theta\right)$ and an estimate of the value function $V\left(s_t ; \theta_v\right)$. 
The RL agent maximizes long-term expected reward in the game in a manner and uses the same mix of $n$-step returns to update the policy and the value function at the same time. 

In text games, actions are strings of tokens.
We do not query the game environments for admissible actions---those that are guaranteed to have an effect. 
However, we do simplify the action space by using templates based on the verb~\cite{ammanabrolu2020graph}.
Templates are composed of interchangeable verbs phrases ($VP$), optionally followed by prepositional phrases ($VP$ $PP$), e.g. $([drink/eat]$ \underline{\hspace*{.3cm}}$)$ and $([apply/ask/put]$ \underline{\hspace*{.3cm}} $[on/about/down]$ \underline{\hspace*{.3cm}}$)$, where the verbs and prepositions within $[.]$ are aliases.
The agent generates actions by first sampling a template and then sampling the word from the game's vocabulary to fill in the blanks.

\label{sec:game_kg}
As the agent explores the game world, we build an internal World KG state representation.
This knowledge graph is distinct from the Story KG.
Following Ammanabrolu {\em et al.}~\shortcite{ammanabrolu2020graph}, 
we consider the process of building a knowledge graph to be a question-answering task.
We fine-tune the ALBERT model~\cite{lan2019albert} on the JerichoQA~\cite{hausknecht2020interactive} dataset, which is specifically designed for question answering in text games. 
This allows the model to answer questions like ``What am I carrying?'' and ``Where am I?''.
We use the answers as a set of candidate vertices $V_t$ for the current step and the questions as a set of relations $R_t$. We then combine $V_t$ and $R_t$ with the game knowledge graph from the previous step $G_{t-1}$ to update the game knowledge graph to $G_t$.
More details can be found in Appendix~\ref{app:qa}.

The ALBERT-QA technique for building the {\em World KG} has been shown to improve RL agent performance because it is trained to the particulars of text games~\cite{ammanabrolu2020graph}.
The {\em Story KG}, on the other hand, is constructed using the VerbAtlas SRL model because exemplar stories draw from a different text distribution---they are expected to be less verbose about world details, intentionally leaving more implicit.
Further, the Story KG only needs to extract a few key details to bias the agent toward certain actions and locations in the environment.

Putting it all together, 
at each step of training, a total score $R_{t}$ and an natural language observation $o_t$ is received from the game environment---consisting of $\left(o_{t_{\text{desc}}}, o_{t_{\text{game}}}, o_{t_{\text{inv}}}, a_{t-1}\right)$ corresponding to the room description, game feedback, inventory, and previous action.
An example is depicted in the left side of Figure~\ref{fig:pipeline}.
As introduced in Section~\ref{sec:game_kg}, the World KG $G_t$ at time step $t$ is also updated.
Each component of $o_t \in \mathbb{R}^{d_{o} \times c}$ is processed using a GRU-based encoder to obtain $\mathbf{o}_{t} $ and World KG, $G_t$, is processed via Graph Attention Networks 
(GATs) \cite{velivckovic2017graph} 
followed by a linear layer to get the graph representation $\mathbf{g}_{\mathbf{t}} \in \mathbb{R}^{d_{o}}$;
 $c$ is the number of $\boldsymbol{o}_{\mathrm{t}}$'s components.
Then we calculate the attention $\alpha$ between $\boldsymbol{o}_{\mathrm{t}}$ and $\mathbf{g}_{\mathbf{t}}$,
\begin{gather}
\boldsymbol{\alpha}_{}=\operatorname{softmax}\left(\boldsymbol{W}_{\text{l}} \boldsymbol{h}_{}+\boldsymbol{b}_{\text{l}}\right)
\\
\boldsymbol{h}_{}=\tanh \left(\boldsymbol{W}_{\mathrm{o}} \boldsymbol{o}_{\text {t}} \oplus\left(\boldsymbol{W}_{\mathrm{g}} \mathbf{g}_{\text {t}}+\boldsymbol{b}_{\mathrm{g}}\right)\right)
\end{gather}
where $\oplus$ denotes the addition of a matrix and a vector and $\odot$ denotes dot-product.
$\boldsymbol{W}_{\mathrm{l}} \in \mathbb{R}^{d_{o} \times d_{o}}$, $\boldsymbol{W}_{\mathrm{g}} \in \mathbb{R}^{d_{o} \times d_{o}}$, $\boldsymbol{W}_{\mathrm{o}} \in \mathbb{R}^{d_{o} \times d_{o}}$ are weights and $\boldsymbol{b}_{\mathrm{l}} \in \mathbb{R}^{d_{o}}$, $\boldsymbol{b}_{\mathrm{o}} \in \mathbb{R}^{d_{o}}$ are biases.
Finally, the overall representation vector $\mathbf{v}_{\mathbf{t}}$ is updated by
$    
\mathbf{v}_{\mathbf{t}} = \mathbf{g}_{\mathbf{t}} + \sum_{i}^{c} \boldsymbol{\alpha}_{i} \odot \mathbf{o}_{\mathrm{t}, i}
$

\subsection{Rewarding the Agent}
\label{sec:intrinsic}

After obtaining the overall representation $\mathbf{v}_{\mathbf{t}}$ above, we incorporate two intrinsic rewards into the RL agent's training, in order to motivate it to act in a manner that to be more closely resemble the implicit world of the story. 

The \textit{KG intrinsic reward} is determined by comparing the similarity between the agent's World KG and Story KG. 
This reflects how closely the agent's actions align the actual world with the world that should exist according to the story.
When new edges are added to the World KG, $\mathbf{G}_t$, we verify if the corresponding triples already exist in the Story KG. Each newly discovered triple results in a positive intrinsic reward $r_t^\mathrm{s} = n \times \rho > 0$, where $n$ is the number of same triples. 
An example can be found in Figure~\ref{fig:pipeline} (the red edges in the knowledge graphs).
For example, the agent performs an action that resembles the implicit world of the story, such as ``drink coffee'', a new edge, $\langle$\texttt{You}, \texttt{DRINK}, \texttt{coffee}$\rangle$, will be added to the World KG.
This triple is identified as being identical to one triple in the Story KG, so the \textit{KG intrinsic reward} $r_t^\mathrm{s} = 1 \times \rho$ at this step.
The edge $\langle$\texttt{You}, \texttt{in}, \texttt{kitchen}$\rangle$ (the green edge in the knowledge graphs) is not considered here as we only take into account new edges added during the current round of the game.

Inspired by Ammanabrolu {\em et al.}~\shortcite{ammanabrolu2020avoid}, we also encourage the agent to explore more locations and scenarios by defining a \textit{exploration  intrinsic reward} 
$
    r_{t}^\mathrm{e}=\Delta\left(\mathbf{G}_{\text {global }}-\mathbf{G}_t\right) 
$,
where $\mathbf{G}_{\text {global }}=\bigcup_{i=1}^{t-1} \mathbf{G}_i$ is the set of all edges that the agent has ever had in its knowledge graph.
When the agent learns more information about the world, it will expand the size of its World KG, increasing the likelihood of reward and success.

The overall intrinsic reward received at time step $t$ is:
\begin{equation}
    r'_t = r_{t} + \alpha \times r_{t}^\mathrm{s} + \beta \times r_{t}^\mathrm{e} 
\end{equation}
where $\alpha$ and $\beta$ are scaling factors; $r_t$ is the game score; $r'_t$ is the reward provided to the agent on time step.
The rest of the training methodology is unchanged from Ammanabrolu {\em et al.}~\shortcite{ammanabrolu2020avoid}.

\section{Experiments}
\label{sec:exp}

We conducted experiments in three phases:
\begin{enumerate}
[noitemsep,topsep=0pt,itemsep=0pt, leftmargin=*]
\item 
We train agents to play text games in which the agent must successfully navigate some normative everyday routines.
Agents are provided with different exemplar stories about the tasks.
In this set of experiments, we evaluate whether Story Shaping facilitates the expression of commonsense and social norm knowledge (Section~\ref{sec:exp-human}).

\item 
We train agents to play an open-ended role-playing game, using different exemplar quests to create different personas.
In this set of experiments, we evaluate whether Story Shaping is capable of shaping the agent's behaviors in a way recognizable to humans.
It also demonstrates that Story Shaping can learn {\rm trope} knowledge, which is knowledge particular to different storytelling conventions (Section~\ref{sec:exp-setting}).

\item
We train agents to play games in which the exemplar stories differ from the operating environment either by referencing objects that do not exist or actions that cannot be performed. 
This set of experiments evaluates whether Story Shaping is robust to different assumptions between the provider of the story and the agent's environment. 
These differences can arise for a number of reasons, one of which being that the story is generated by another system, such as ChatGPT~\cite{chatgpt} (Section~\ref{sec:chatgpt_exp}).

\end{enumerate}

\paragraph{Baselines}
We implement the RL agents below with and without Story Shaping:
\begin{itemize}
    \item \textbf{Q*BERT}~\cite{ammanabrolu2020avoid}, a state-of-the-art RL agent for text games designed to work with Jericho games and TextWorld.
    It constructs its World KG by answering questions with ALBERT-QA.
    \item \textbf{KG-A2C}~\cite{ammanabrolu2020graph}, which uses 
    Stanford's Open Information Extraction (OpenIE)~\cite{angeli2015leveraging} to build its World KG because ALBERT-QA is tuned on the JerichoQA dataset. 
    KG-A2C is used for LIGHT experiments.
\end{itemize}

\subsection{Commonsense and Social Norm Knowledge }
\label{sec:exp-human}
We first seek to understand whether Story Shaping enhances the expression of common sense and social norm knowledge in Reinforcement Learning agents.

\paragraph{Games.}
We implement three slice-of-life text games on two game platforms:
\begin{itemize}
    \item \textit{9:05}: a game in which the agent must successfully navigate the normative routine of getting ready to work, implemented in Jericho~\cite{hausknecht2020interactive}, a framework for interacting with text games, as the interface connecting learning agents with interactive fiction games. 

    \item 
    \textit{Shopping}: a game in which the agent must successfully purchase the clothes, developed in TextWorld~\cite{cote2018textworld},
    an open-source, extensible engine that both generates and simulates text games. 

    \item 
    \textit{See Doctor}: a game in which the agent gets sick and must seek medical treatment, also developed in 
    TextWorld.
\end{itemize}
Details for {\em 9:05} can be found in Appendix~\ref{app:905}.
Details for {\em Shopping} and {\em See Doctor} can be found in Appendix~\ref{app:textworld}.

\paragraph{Training.}
For each game, we 
train two agents. 
{\em Q*BERT-S} uses Story Shaping with stories written by humans.
The baseline, {\em Q*BERT} is the same agent but without Story Shaping.
We evaluate these two trained agents by running test games over 20 random seeds.
Training details are in Appendix~\ref{app:agents_training}.

\paragraph{Automatic Evaluation.}
We automatically evaluate the agents' expression of common sense and social norm knowledge  by:
\begin{itemize}
    \item \textbf{Win rate}: the winning rate of trained agents on test games over 20 random seeds.
    \item \textbf{Avg steps}: the average number of steps that each agent takes to win the game. 
    The game will automatically end over $50$ steps.
    \item \textbf{Avg Commonsense score}:
    the total intrinsic reward accrued over a testing trial; 
    a higher score indicates the agent takes more actions that express commonsense and social norm knowledge.
    Details about this test game environment can be found in Appendix~\ref{app:test-game}.
    
    \item \textbf{Avg game score}:
    the average score of each agent on the test games, which reflects how far toward the win condition the agent made it, irrespective of {\em how} the agent reached the farthest point. 
\end{itemize}

\noindent
Results are shown in Table~\ref{tab:auto}.
The win rates and the average game scores are identical between agents with and without Story Shaping. 
However, our agent's ``Avg Commonsense Score'' is significantly higher than the baseline agent's, indicating that the Story Shaping agent demonstrates superior ability in expressing common sense and social norm knowledge.
The larger "Avg Steps" value for our agent also suggests that it takes more actions before winning the game, which further highlights that it is not seeking the shortest possible trajectory.

\begin{table}[t]
\footnotesize
\centering
\setlength\tabcolsep{1.7pt} %
\begin{tabular}{c|l||c|r|r|r|r|r}
    \toprule
    \multirow{3}{*}{\textbf{Game}}
    &\multirow{3}{*}{\textbf{Agents}} 
     &\textbf{Win }&
    \multirow{2}{*}{\textbf{Avg} \hspace{0.05cm}}&
    \textbf{Avg \hspace{0.05cm}} & \textbf{Max \hspace{0.05cm}}&
    \textbf{Avg \hspace{0.05cm}} & \textbf{Max \hspace{0.05cm}}\\

    && \textbf{Rate} & \multirow{2}{*}{\textbf{Steps}} 
    & \textbf{CS \hspace{0.1cm}}
    &\textbf{CS \hspace{0.1cm}}
    &\textbf{Game}&\textbf{Game}\\
    &&\%&&\textbf{Score}&\textbf{Score}&\textbf{Score}&\textbf{Score}\\
    \midrule

    \multirow{2}{*}{\texttt{9:05}}
    &\sysnameq{} &
      100&16.30&\textbf{3.90}&\multirow{2}{*}{4}&5.00&\multirow{2}{*}{5}\\ 
    
    &Q*BERT&
      100&7.25&0.40&&5.00\\
    \midrule

    \multirow{2}{*}{\texttt{Shopping}}
    &\sysnameq{} &100&12.35&\textbf{3.70}&\multirow{2}{*}{4}&5.00&\multirow{2}{*}{5}\\ 
    
    &Q*BERT&100&6.30&0.90&&5.00\\
    \midrule

    \multirow{2}{*}{\texttt{Doctor}}
    &\sysnameq{} 
    &95&19.15&\textbf{6.70}&\multirow{2}{*}{8}&4.75&\multirow{2}{*}{5}\\ 
    
    &Q*BERT&95&14.30&0.70&&4.75&\\

    \bottomrule
\end{tabular}

\caption{Automatic evaluation results across 20 independent runs comparing \sysnameq{} to baseline Q*BERT
Each system is trained under the same game environment. 
}
\label{tab:auto}
\end{table}

\begin{table}[t]
\footnotesize
\centering
\setlength\tabcolsep{0.6pt} %
\begin{tabular}{c||lrc|ccc}
    \toprule
    \multirow{2}{*}{\textbf{Game}}
    &\multicolumn{3}{c|}{\textbf{Commonsense/Social Norm}}
     &\multicolumn{3}{c}{\textbf{Understanding}}\\
    
    & Shaped \%
    & Base \%    
    & Tie \%
    & Shaped \%
    & Base \%
    & Tie \%\\
    
    \midrule

    \texttt{9:05}
      & \textbf{63.63*}
      & 9.09
      & 27.27
      & 36.36
      & \textbf{45.45}
      & 18.18\\ 

    \texttt{Shopping}
    & \textbf{66.67**}
      & 8.33
      & 25.00
      & \textbf{41.67}
      & 33.33
      & 25.00\\ 

    \texttt{Doctor}
    & \textbf{53.85}
      & 23.08
      & 23.08
      & \textbf{46.15}
      & 38.46
      & 15.38\\ 
\bottomrule
\end{tabular}
\caption{Human evaluation results showing the percentage of participants who preferred Story Shaped \sysnameq{} to baseline Q*BERT, or
thought the systems were equal. 
Each system is trained under the same game environment. 
* indicates human evaluation results are significant at $p < 0.05$ confidence level; ** at $p < 0.01$ using a Wilcoxan sign test on win-lose pairs.
}
\label{tab:human}
\end{table}

\paragraph{Human Evaluation.}

We recruited $30$ participants using the Cloud Research platform and Amazon Mechanical Turk \cite{litman2017turkprime}.
We screened for participants that were generally not familiar with text games.
Each participant reads the winning goal of a randomly chosen game.
Then they read a pair of game transcript which played by \sysnameq {} and Q*BERT, specifically. 
Each transcript includes game observations and the corresponding actions. 
Then they are given the following metrics and asked to choose which  game transcript they prefer for that metric:
\begin{itemize}
    \item This sequence of actions expresses more common sense thinking (with social norm knowledge) on the action choice.
    \item This sequence of actions makes you understand why the agent takes these actions given what you know about the goal.
\end{itemize}
Participants had to provide detailed explanations for their choices in each comparison, using at least 50 characters of free text. 
We manually reviewed responses and discarded data from those who did not provide detailed explanations.
Each game trajectory pair is evaluated by a minimum of $10$ participants.
Our study was approved by our Institutional Review Board, and we payed participants the equivalent of $\$15$/hr. 
Example of the human evaluation survey can be found in Appendix~\ref{app:human_exp1}

Table~\ref{tab:human} shows the percentage of times
stories from each system are preferred for each metric. 
In the same game environment, \sysnameq{} performs significantly better than Q*BERT on the dimension of ``Common and Social Norm Sense''.
We can conclude that Story Shaping facilitates the expression of commonsense and social norm knowledge of the trained RL agents significantly.
On the dimension of ``Understanding'', 
we would expect Story Shaped agents to be no less understandable than the baseline, indicating that longer trajectories are not random.
\sysnameq{} achieves comparable results with Q*BERT. 
Participants who favored Q*BERT mentioned that they found the shorter game paths easier to comprehend. As seen in Table~\ref{tab:auto}, Q*BERT takes fewer steps to complete the game, making it more straightforward for human participants to follow.
Our system is thus shown to improve the expression of commonsense and social norm knowledge of agents while preserving comprehensibility.

\subsection{Persona Understanding}
\label{sec:exp-setting}

We assess whether Story Shaping has the ability to shape the agent's behaviors in a way that is identifiable to humans.
We develop a role-playing game in the LIGHT~\cite{urbanek2019learning} environment, which is a large-scale fantasy text adventure game research platform for training agents that can both talk and act, interacting either with other models or with humans.
We provided different exemplar quests for four personas: \texttt{thief}, \texttt{bum}, \texttt{adventurer} and \texttt{thug}.
We provide a LIGHT world that provides a rich set of locations and objects for all personas to make use of, or ignore.
Details about this game can be found in Appendix~\ref{app:games_light}.
In these experiments, we compare two KG-A2C agents utilizing Story Shaping but with different human-written stories.

We recruited an additional $29$ participants. 
Participants will read a winning goal and be told the agent's persona (thief, adventurer, thug, bum), then read two game transcripts played by versions of KG-A2C with Story shaping. One version is trained using a story about the given persona, the other using a random story selected from the remaining personas.
Participants were asked to indicate which of the two agents had the given persona.
At least $10$ participants evaluate each game.
More details about this human evaluation survey can be found in Appendix~\ref{app:human_exp2}.

\begin{table}[t]
\footnotesize
\centering
\setlength\tabcolsep{3pt} %
\begin{tabular}{c|c||lcc|c}
    \toprule
     \multirow{2}{*}{\rotatebox[origin=c]{90}{\tiny{\textbf{Game}}}}& \textbf{Given}
     &\multicolumn{3}{c||}{\textbf{Participant Choice}}&\textbf{Win}\\
    & \textbf{Persona}&Correct\%& Incorrect\% & Can't tell\% &\textbf{Rate}\\
    \midrule
    \multirow{4}{*}{\rotatebox[origin=c]{90}{\texttt{LIGHT}}}
    & \texttt{Thief}& 
    \textbf{58.06*}&19.35&22.58 & 100
      \\ 
    & \texttt{Adventurer}& 
    \textbf{62.50*}
    & 15.62 & 21.88 & 100
      \\ 
    & \texttt{Thug}& 
    \textbf{72.73**}&12.12&15.15 & 100
      \\ 
    & \texttt{Bum}& 
    \textbf{64.71*}  & 23.53 & 11.76
    & 100
    \\
    \bottomrule
\end{tabular}

\caption{The percentage of participants who preferred the agent utilizing Story Shaping with the specific story of the given persona, the agent using Story Shaping with a story from a different persona, or believed the systems were equivalent when the game goal and persona were provided to human participants.
``Win Rate'' is the winning rate of trained agents using Story Shaping with the specific story of the given persona on test games over 20 random seeds.
The symbols (* and **) used for indicating significance in as in Table~\ref{tab:human}.
}
\label{tab:human_setting}
\end{table}
\begin{table}[t]
\footnotesize
\centering
\setlength\tabcolsep{2pt} %
\begin{tabular}{c|c||lcr|ccr}
    \toprule
    \multirow{2}{*}{\textbf{Game}}
     &\multirow{2}{*}{\textbf{Persona}}
     &\multicolumn{3}{c|}{\textbf{Common/Social}}
     &\multicolumn{3}{c}{\textbf{Understanding}} 
     \\
    & &GPT\%& Base\% & Tie\%&GPT\%& Base\% & Tie\%
    \\

    \midrule

    \multirow{1}{*}{\texttt{9:05}}
    & -& \textbf{71.43*}&14.29&14.29&
    \textbf{57.14}&42.86&0.00 
      \\ 
    \midrule
    \multirow{1}{*}{\texttt{Shop.}}
    & -& \textbf{62.50*} & 25.00 & 12.50
    & \textbf{37.50} &25.00& \textbf{37.50}
      \\

    \midrule

    \multirow{3}{*}{\texttt{LIGHT}}
    & \texttt{Thief}& 
    \textbf{57.14}&28.57&14.29
    & \textbf{42.86}& 28.57 & 28.57
      \\ %

    & \texttt{Adv}& \textbf{83.33**} & 16.67 & 0.00
    & \textbf{50.00} & 16.67 & 33.33
      \\ %

    & \texttt{Thug}& 
    \textbf{64.71*} & 11.76 & 23.53
& \textbf{52.94} & 35.29 & 11.76
      \\ 
    \bottomrule
\end{tabular}

\caption{The percentage of participants who favored the agent with Story Shaping using the ChatGPT-generated story over the baseline, or deemed the systems as indistinguishable.
The symbols (* and **) are used for indicating significance as in Table~\ref{tab:human}.
}
\label{tab:tab3}
\end{table}

Table~\ref{tab:human_setting} displays the percentage of participants who chose the game transcript generated by the agent with the exemplar story for the given persona, versus a randomly chosen exemplar story.
The percentage reflects the effectiveness of Story Shaping in shaping the agent's behaviors in a way that is recognizable to humans.
The results indicate that RL agent with Story Shaping is able to comprehend exemplar quests for different personas and generate actions that align with the given persona.
Additionally, all the versions of KG-A2C with Story Shaping attain a win rate of 100\%, regardless of which exemplar story.
It demonstrates the capability of Story Shaping in shaping the agent's behaviors in a way that is recognizable to humans while at the same time do not
lose any performance.

\subsection{Robustness}
\label{sec:chatgpt_exp}
Our final experiment
investigates the robustness of Story Shaping to variations in assumptions between the exemplar story provider and the agent's environment.
Instead of using stories designed by humans based on their understanding of the environment, we use automatically generated stories by ChatGPT.
Notably ChatGPT has no familiarity with the particular game world and 
these exemplar stories diverge substantially from the agent's game environment, such as referencing objects that do not exist or providing actions that are not available.

We prompt ChatGPT with a description of the first game state, as well as an optional character personality to obtain the guiding story that can then be converted to a knowledge-graph triple format to be used in guiding the agent. 
We then proceed to replicate the training methodology outlined in Section~\ref{sec:exp-human}.
More examples of the produced descriptions and resulting triples are shown in Appendix~\ref{app:chatgpt}.
We follow the same evaluation process and recruited $34$ participants on a crowd sourcing platform to answer the same questions with Section~\ref{sec:exp-human}.
A minimum of $10$ participants evaluate every pair of game trajectories.

Table~\ref{tab:tab3} shows the preference percentage for the stories from each system in each metric.
The high preference percentage shows that Story Shaping with the automatically generated story can also guide RL agents to express more common sense and social norm knowledge, even though
some information in the automatically generated story's knowledge graph is unattainable.
For example, in Table~\ref{tab:ex_actions}, the  ChatGPT generated exemplar story involves an entity that does not exist in the game---``directory listings''.  %
Despite the mismatch, our technique allows for the flexibility needed for our agent to successfully complete tasks by utilizing other elements of the exemplar story and using trial-and-error to fill in the rest. Agents have a 100\% game completion rate in all the test game environments.

\begin{table}[t]
\centering
\footnotesize
\setlength\tabcolsep{0.1pt} %
\begin{tabular}{p{0.99\linewidth}}
\toprule
\textbf{Game: \texttt{shopping}}  \\
\textbf{First observation:}
I am in front of a mall.
A cafe lies to the south, while
a way to the east leads to the mall.\\
\textbf{Goal: BUY CLOTHES}. \\
\midrule
\midrule
\textbf{Exemplar story from ChatGPT:}\\
To buy clothes, you \underline{\textit{enter the mall}} and navigate to the store that
sells the clothing you are interested.
Many malls have \uwave{directory listings} near the entrances, which
can help you \textit{find the specific stores} you are looking for.
Once locating the store, you can \underline{\textit{browse}} and \underline{\textit{try on the clothing}}, and then \underline{\textit{make a purchase}} at the register.\\
\midrule
\midrule
\textbf{Example action sequence by \sysnameq tory Shaping:}\\
Location: \textit{Street} \hspace{0.65cm} Action: \textit{go east}\\
Location: \textit{Mall} \hspace{0.77cm}  Action: \textit{go north}\\
Location: \textit{Store} \hspace{0.7cm}  Action: \textit{\underline{examine clothes}; \underline{try clothes}}\\
\hspace{2.75cm}  Action: \textit{give money to staff}\\
\hspace{2.75cm}  Action: \textit{buy clothes; take clothes}\\

\midrule
\textbf{Example action sequence by Q*BERT:}\\
Location: \textit{Street} \hspace{0.65cm} Action: \textit{go east}\\
Location: \textit{Mall} \hspace{0.77cm}  Action: \textit{go north}\\
Location: \textit{Store} \hspace{0.7cm}  Action: \textit{give money to staff}\\
\hspace{2.75cm}  Action: \textit{buy clothes; take clothes}\\

\bottomrule

\end{tabular}
\caption{Example action sequence in the game \texttt{shopping}. 
Words \underline{underlined} indicate actions that can be taken in the game environment.
Words with a wavy \uwave{underline} indicate actions that are invalid or not allowed in the game.
}
\label{tab:ex_actions}
\end{table}

\section{Conclusions}

Story Shaping is a straightforward approach to the challenge of reward design where one wishes to not only reward an agent for completing a task, but reward the agent for {\em how} it accomplishes the task.
This might mean aligning an agent's behavior with human preferences and expectations, teaching the agent commonsense reasoning and social norms, or shaping character personas in a game.

Our technique allows one to provide a high-level exemplar story from which the agent automatically extracts knowledge about important objects, locations, and actions.
It then self-rewards when these objects, and locations are encountered, and actions are performed.
Because stories are high-level abstractions, Story Shaping can fill in missing details and is robust to situations where the story cannot be exacted as given.
We have shown that Story Shaping has a significant and human-observable impact on agent behavior without compromising task completion.

\clearpage
\bibliographystyle{named}
\bibliography{anthology,custom}

\clearpage
\appendix
\section{Implementation Details}

\subsection{Semantic Role Labeling Using VerbAtlas}
\label{app:verbatlas-srl}
The SRL model automatically identifies and labels the argument structures of stories.
For example, it extracts \texttt{\text{`verbatlas'}: \text{`EXIST\_LIVE'},  \text{`args\_words'}: \{\text{`Theme'}: \text{`Jenny'}, \text{`Attribute'}: \text{`Georgia'}\}} from ``\textit{Jenny lived in Georgia}''.
Verbs in the story will be represented as the VerbAtlas frame.
For example, \texttt{``live''} is represented as \texttt{``EXIST\_LIVE''}.

We use a fine-tuned transformer model for semantic role labeling (SRL), which is a BERT \cite{devlin2019bert} model with a linear classification layer trained on the Ontonotes 5.0 dataset to predict PropBank\cite{palmer2005proposition} SRL. This model, proposed by Shi~\shortcite{shi2019simple}, is currently the state-of-the-art for English SRL.  
We use an open-source implementation \footnote{\href{https://github.com/Riccorl/transformer-srl}{https://github.com/Riccorl/transformer-srl}}, which is based on the official AllenNLP BERT-SRL model \footnote{\href{https://demo.allennlp.org/semantic-role-labeling}{https://demo.allennlp.org/semantic-role-labeling}}. Trained with the following hyperparameters:

\begin{compactitem}
    \item Batch size: 32
    \item Dropout for the input embeddings: 0.1
    \item Learning rate: $5e^{-5}$
    \item Optimizer: Adam
    \item Total Epochs: 15
\end{compactitem}

Then, we use the mappings from Propbank frames to VerbAtlas \cite{di2019verbatlas} classes to return the correct corresponding VerbAtlas classes instead of Propbank's \cite{palmer2005proposition}. We can directly map VerbAtlas classes to PropBank frames because for every VerbAtlas class, there is only one PropBank frame. This allows us to use the rich content from VerbAtlas with the same model that was initially trained to predict PropBank.

\subsection{Knowledge Graph Representation QA Model}
\label{app:qa}

The question answering network based on ALBERT~\cite{lan2019albert} uses the hyperparameters listed in the original paper. These hyperparameters have been shown to work well on the SQuAD 2.0~\cite{rajpurkar-etal-2018-know} dataset.
We did not do any additional tuning of the hyperparameters.
\begin{center}
\begin{tabular}{l|r}
    \textbf{Parameters} & \textbf{Value}\\ 
    \hline 
    batch size & 8\\
    learning rate & 3e-5\\
    max seq len & 512\\
    doc stride & 128\\
    warmup steps & 814\\
    max steps & 8144\\
    gradient accumulation steps & 24\\
 \end{tabular}
\label{tab:qbertParams}
\end{center}

\subsection{RL agents with Story Shaping}
\label{app:agents_training}

Further details of what is found in Figure~\ref{fig:pipeline}.
The sequential action decoder consists two GRUs that are linked together as seen in Ammanabrolu~\shortcite{ammanabrolu2020graph}.
The first GRU decodes an action template and the second decodes objects that can be filled into the template.
These objects are constrained by a {\em graph mask}, i.e. the decoder is only allowed to select entities that are already present in the knowledge graph.

Same with Ammanabrolu~\shortcite{ammanabrolu2020graph}, the loss consists of template loss, object loss, value loss, actor loss and entropy loss.
The template loss given a particular state and current network parameters
is applied to the decoder.
Similarly, the object loss is applied across the decoder is calculated by summing cross-entropy loss from all the object decoding steps.
Entropy loss over the valid actions, is designed to prevent the agent from prematurely
converging on a trajectory.
The following hyperparameters are taken from the original paper and known to work well on text games.

\begin{center}
\begin{tabular}{l|r}
    \textbf{Parameters} & \textbf{Value}\\ 
    \hline 
    discount factor & 0.9 \\
    entropy coefficient & 0.03\\
    value coefficient & 9\\
    template coefficient & 3\\
    object coefficient & 9\\
 \end{tabular}
\label{tab:qbertParams}
\end{center}

\subsection{Test Game with Human-likeness Score}
\label{app:test-game}
In order to automatically evaluate the expression of commonsense and social norm knowledge of the trained RL agents, 
we develop a test game mode for each text game.
During training, the player can only earn a score if they win the game. 
However, when evaluating the RL agents, we test both the baseline and our agents in a test game that has the same map and actions as the training game, but a different scoring system. 
Specifically, the test game environment not only gives a score when the player wins the game, but also gives scores when the agent reaches specific states. 
For example, in the game \texttt{9:05}, the test game environment will give a score of 2 when the agent eats a pop-tart in the kitchen, whereas in the training game, this action would not earn any points.
Our test game mode allows us to automatically evaluate the human-like qualities of our trained RL agents. We believe that a higher score in this game mode indicates a greater degree of the expression of commonsense and social norm knowledge of the trained RL agents.

\newpage
\section{Game Design}
\label{app:games}
\subsection{Jericho Game}
\label{app:905}
\paragraph{9:05} 
is a text adventure game by Adam Cadre. The game is designed for a casual audience, including those new to the genre. 
It is a short and simple game that follows a branching narrative in which a man wakes up and receives a call telling him to go to work. 
Our agent can be tested effectively in this game due to the numerous optional commonsense actions available to the player. In the game, the player has the option to leave their home and go directly to work in order to win the game immediately, 
or they can choose to take some time to get ready by changing clothes, taking a shower, and having a toast before starting their day. 
All of these actions are optional and do not affect the outcome of the game.

\textit{Exemplar story} we use for Section~\ref{sec:exp-human}: ``
Upon waking in the morning, I start my day with a Pop-Tart breakfast, followed by a shower before commencing work."
''

We designed a game with optional \textit{CS scores} for evaluating the human-likeness of our agents.
CS scores designed are as follows,
\begin{itemize}
    \item Score 2: \texttt{shower} is used.
    \item Score 2: \texttt{Pop-Tart} is consumed.
\end{itemize}

\subsection{TextWorld}
\label{app:textworld}
We designed two games in TextWorld \cite{cote2018textworld} text game engine.

\begin{figure}
    \centering
\includegraphics{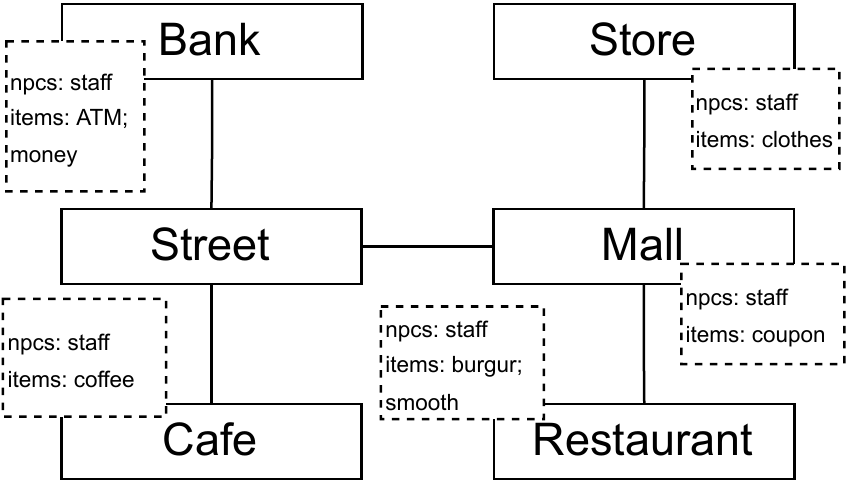}
    \caption{Layout of game ``shopping''.}
    \label{fig:shopping_map}
\end{figure}

\paragraph{Shopping} 
is a game where the goal is to purchase clothing at a mall. 
The player starts out on the street and has the option to stop at various cafes and restaurants on the way to the mall. They may also encounter NPC characters who can provide information on obtaining coupons. 
The player has multiple routes they can take to reach the clothing stores and can choose to engage in optional activities such as purchasing a coffee or using coupons. 
The layout of the game is shown in Fig.\ref{fig:shopping_map}.

The game score is only obtained when the player finish the game.
When the player has ``\texttt{clothes}''  in his inventory, the game ends with a game score: $5$.

\textit{Exemplar story} for experiments in Section~\ref{sec:exp-human} is: ``
To save money, I need to obtain a coupon. Once I have tried on the clothes, I will purchase them.
''

We designed a game with optional \textit{CS scores} for evaluating whether Story Shaping facilitates the expression of commonsense and social norm knowledge.
CS scores designed for Section~\ref{sec:exp-human} are as follows,
\begin{itemize}
    \item Score 2: \texttt{coupon} is applied.
    \item Score 2: \texttt{clothes} is tried.
\end{itemize}

\paragraph{See Doctor}
is a game where the objective is to obtain the medicine. 
The player starts at home with a cup of hot water and money. 
The player can make hot soup, buy hot coffee, visit the doctor at the hospital, or go to a drug store to get the medicine.
The player has multiple paths they can take to obtain the medicine, and they can also take various optional actions such as buying a coffee or drinking hot water.

\begin{figure*}[tbh!]
    \centering
    \includegraphics{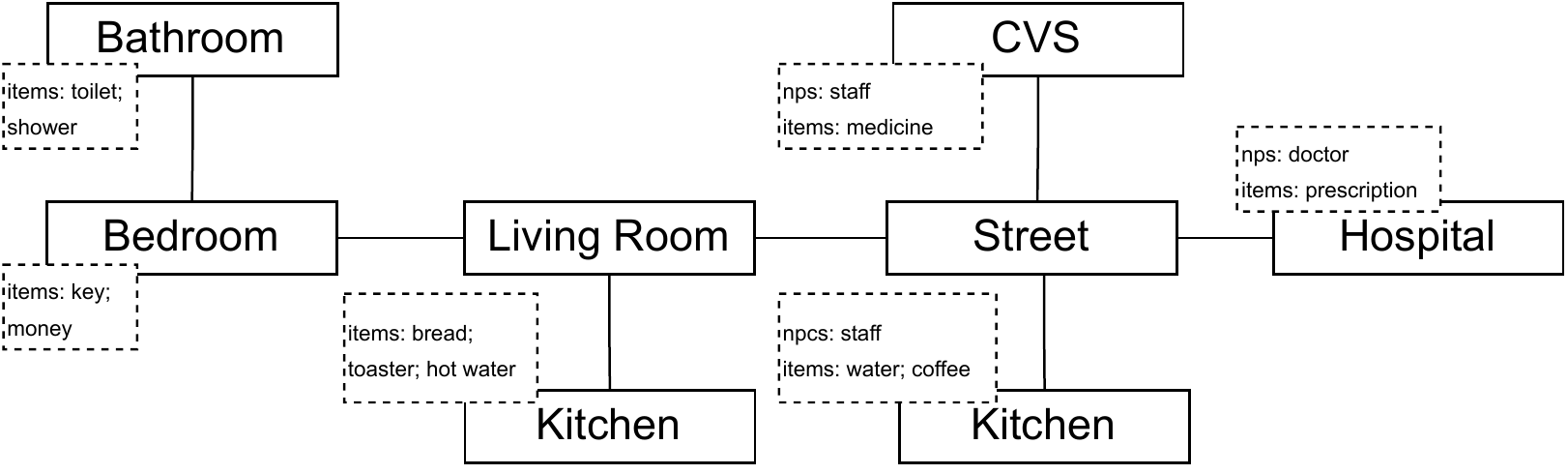}
    \caption{Layout of game ``see doctor''.}
    \label{fig:seedoc_map}
\end{figure*}

The game score is only obtained when the player finish the game.
When the player has ``medicine'' in his inventory, the game ends with a game score: $5$.
You can see the layout of this game in Fig.\ref{fig:seedoc_map}

\textit{Exemplar Story} for experiments in Section~\ref{sec:exp-human} is: ``
I caught a cold and drank hot water, but it didn't help after taking a shower. I went to the hospital to see the doctor and get a prescription to buy medicine.
''

We designed a game with optional \textit{CS scores} for evaluating whether Story
Shaping facilitates the expression of commonsense and social norm knowledge.
The
\textit{CS scores} designed are as follows,
\begin{itemize}
    \item Score 2: \texttt{water} is consumed.
    \item Score 2: \texttt{shower} is used.
    \item Score 2: \texttt{doctor} is seen.
    \item Score 2: \texttt{prescription} is in inventory.
\end{itemize}

\subsection{LIGHT}
\label{app:games_light}
LIGHT~\cite{urbanek2019learning} is a text adventure game research platform for training agents that can both talk and act. 
For our experiments, we developed a large map to serve as a sandbox. 
Our agent can be tested effectively in this game due to the large range of actions available to the player. 
These actions include interacting with objects and NPC characters and navigating through the map. 
All actions are optional, and the game is terminated once the agent reaches a designated goal location.

Urbanek~\shortcite{urbanek2019learning}~utilized generative and retrieval models to allow NPC characters to exchange dialogue with players. 
However, we do not leverage these models and restrict the agent's interactions with NPC characters to actions as the addition of dialogue adds unnecessary complexity that is not needed for our experiments.

\begin{figure*}[tbh!]
    \centering
    \includegraphics[width=\textwidth]{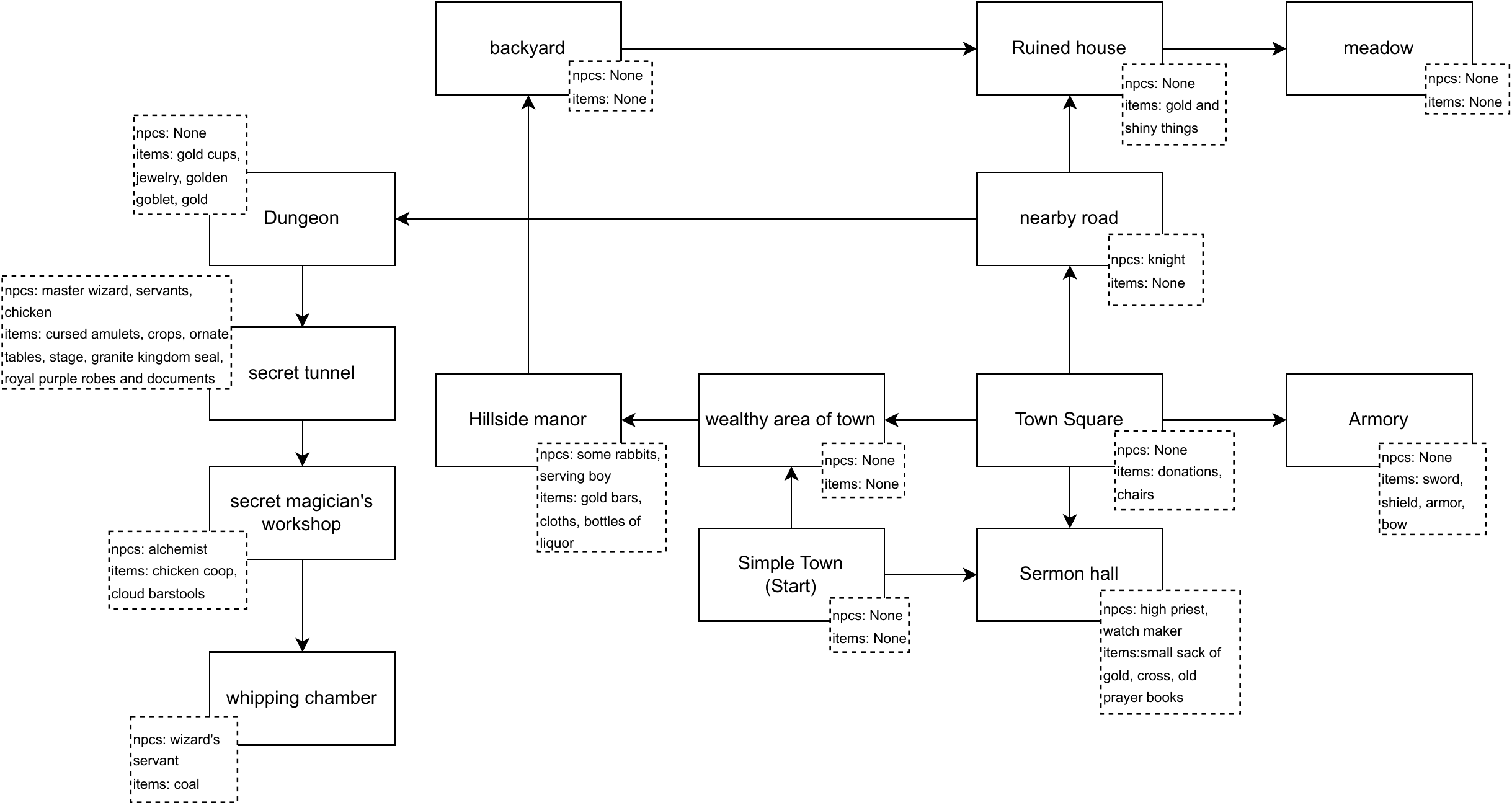}
    \caption{Layout of LIGHT Map used in testing
    }
    \label{fig:light_map}
\end{figure*}

We design a  role-playing game, ``gold'', where the player can play different personas, collect items, and then finally ``go to Meadow''. The player starts out in the Simple Town and has the option to visit various locations in the town such as the wealthy area of town, the Sermon Hall or the Armory and collect different items at these locations. The player has multiple routes they can take to reach the Meadow and can choose to engage in optional activities such as visiting the Armory, or picking up old prayer books in the Sermon Hall. The layout of the game is shown in Fig. \ref{fig:light_map}.

The game score is only obtained when the player finish the game.
When the player enters the \textit{Meadow}, the game ends with a game score: $5$.

Additionally, we also demonstrate that our system is capable of generalizing to other guiding narratives, so long as they are capable of being navigated within the environment. To this end, we provide a fourth guiding narrative with a different goal that is still achievable within the game.

\textit{Exemplar story} of persona \texttt{Adventurer}: ``
I am a brave Adventurer. I know there is a Dungeon with valuable treasure. I go to the armory and get a sword, a shield, armor and a bow to defend myself. Once I am properly equipped, I go to the Dungeon. I get the gold, jewelry, gold cups and a golden goblet. I then leave through the Meadow.
''

We designed a persona with optional \textit{CS scores} for evaluating whether Story Shaping facilitates the expression of role-aligned norms within a fantasy environment.
CS scores designed for Section~\ref{sec:exp-setting} are as follows,
\begin{itemize}
    \item Score 10: \texttt{sword} is obtained.
    \item Score 10: \texttt{armor} is obtained.
    \item Score 10: \texttt{shield} is obtained.
    \item Score 10: \texttt{bow} is obtained.
    \item Score 10: \texttt{gold} is obtained.
    \item Score 10: \texttt{jewelry} is obtained.
    \item Score 10: \texttt{gold cups} is obtained.
    \item Score 10: \texttt{golden goblet} is obtained.
\end{itemize}

\textit{Exemplar story} of persona \texttt{Thief}: ``
I am a cowardly Thief. I go to the wealthy area of town to search for valuables. I enter the Hillside Manor and get the gold bars there. I stealthily go to the Sermon Hall, and get the small sack of gold. I then leave through the Meadow.
''

We designed a persona with optional \textit{CS scores} for evaluating whether Story Shaping facilitates the expression of role-aligned norms within a fantasy environment.
CS scores designed for Section~\ref{sec:exp-setting} are as follows,
\begin{itemize}
    \item Score 10: \texttt{gold bars} are obtained.
    \item Score 10: \texttt{small sack of gold} is obtained.
\end{itemize}

\textit{Exemplar story} of persona \texttt{Bum}: ``
I am a lazy bum. I wish to do as little as possible to get some coins and leave. I immediately exit the Simple Town. I only stop at the Town Square to get the donations before leaving through the Meadow.
''

We designed a persona with optional \textit{CS scores} for evaluating whether Story Shaping facilitates the expression of role-aligned norms within a fantasy environment.
CS scores designed for Section~\ref{sec:exp-setting} are as follows,
\begin{itemize}
    \item Score 10: \texttt{donations} are obtained.
\end{itemize}

\textit{Exemplar story} of persona \texttt{Thug}: ``
I am a Thug. I wish to get revenge on the watch maker who scammed me. I find the watchmaker in the Sermon Hall. I hit the watch maker, and he falls to the floor dead. I then leave through the Meadow.
''
We designed a persona with optional \textit{CS scores} for evaluating whether Story Shaping facilitates the expression of role-aligned norms within a fantasy environment.
CS scores designed for Section~\ref{sec:exp-setting} are as follows,
\begin{itemize}
    \item Score 5: \texttt{watch maker} is hit.
\end{itemize}

\newpage
\section{ChatGPT}
\label{app:chatgpt}
ChatGPT is a state-of-the-art language generation model developed by OpenAI. It is based on the GPT (Generative Pre-trained Transformer) architecture, which uses machine learning to generate human-like text. ChatGPT is trained on a large dataset of conversational text and is able to generate responses to prompts in a wide range of contexts, making it a powerful tool for natural language processing tasks such as chatbots, language translation, and text summarization.

We find that with proper prompting, ChatGPT is able to generate exemplar stories we need for Story Shaping, since stories can leave
details about the environment out.
In this section, we introduced the prompt we use for automatically generating stories from ChatGPT.

\subsection{Jericho Game: 905}
\label{app:chatgpt_905}
We use the prompt---``[\texttt{observation}]. What will I do in order to [\texttt{GOAL}]'' for prompting ChatGPT.
In the game ``9:05'',  we use the prompt:``I wake up in the morning. A bathroom lies to the south, while a door to the east leads to the living room. On the end table are a telephone, a wallet and some keys. The phone rings. What will I do before \texttt{going to work}?''
The 3 outputs from ChatGPT are as follows, 
\begin{itemize}
    \item ``I will likely take a shower in the bathroom to the south, get dressed, and check my wallet and keys to make sure I have everything I need for the day. I may also take a cup of coffee before leaving my home to go to work.''
    \item ``Before going to work, it is likely that you would answer the phone to see who is calling. You may also check your wallet and keys to ensure that you have everything you need for the day. You may also use the bathroom before leaving your home.''
    \item Before going to work, I will first answer the phone to see who is calling. 
    Next, I will head to the bathroom to freshen up and brush my teeth. 
    After that, I will consider doing some exercise to start my day off on a healthy note. 
    Finally, I will grab my keys and head out the door to start my day.
\end{itemize}

We convert these generated stories into knowledge graphs with verbAtlas frameworks.
Then the agent is trained using Story Shaping with randomly chosen exemplar stories.

\subsection{TextWorld Game: Shopping}
We use the same prompt with Appendix~\ref{app:chatgpt_905}---``[\texttt{observation}]. What will I do in order to [\texttt{GOAL}]''.
In the game ``shopping'',  we use the prompt:``I am in front of a mall. There is a bank in the north and a cafe in the south. What will I do before \texttt{buy clothes}?''
The outputs from ChatGPT are as follows,
\begin{itemize}
    \item ``In order to buy clothes at the mall, you can enter the mall and navigate to the store or stores that sell the type of clothing you are interested in purchasing. Many malls have directory listings near the entrances, which can help you find the specific stores you are looking for. Once you have located the store, you can browse and try on the clothing, and then make a purchase at the register.''
    \item ``In order to buy clothes, I will first check if I have enough money in my bank account by visiting the bank in the north. If my account has sufficient funds, I will proceed to the mall.
    Once inside the mall, I will have a look around the different clothing stores and see if there are any sales or promotions that interest me. I will also check if there's any specific clothes I was looking for.
    After browsing through the different stores, I will make a decision on which clothes I want to purchase.
    Before leaving the mall, I may want to take a break and grab a coffee or a snack from the cafe in the south.
    Finally, I will make my purchase and head home with my new clothes.
    ''
    \item 
    In order to buy clothes, I will first head to the bank in the north to withdraw some cash or check my account balance, if I am low on cash.
    After that, I will head to the mall, which is in front of me, and enter the mall.
    I will then look around the mall to find the clothing stores. I may ask a mall employee for directions if needed.
    Once I find the clothing stores, I will browse through the different options and try on any clothes that catch my interest.
    Finally, I will make my purchases and head to the cafe in the south for a break and enjoy a cup of coffee.
\end{itemize}

We then convert these generated stories into knowledge graphs and apply Story Shaping techniques for the RL agents.

\subsection{LIGHT}

We use a series of prompts for the LIGHT environment to provide ChatGPT with context about the environment as well as the ultimate goal due to the complexity of the LIGHT environment compared to Jericho or Textworld. Additionally, as ChatGPT is not allowed to talk about violence, it is not capable of generating an output for the role of 'Thug'. For this reason, it is not included. ChatGPT was accessed on 1/7/2023.

To provide context about the environment, we use the prompt---``I am going to describe a game. Please use this description to answer my next question. In the world of LIGHT, there is a Simple Town. There is a Sermon hall to the east. In the Sermon hall are a "high priest", a "watch maker", a "small sack of gold", a "cross" and "old prayer books". The wealthy area of town is to the north. In the wealthy area of town there is a Hillside manor. In the Hillside manor, there are some "rabbits", a "serving boy", "gold bars", "cloths" and "bottles of liquor". To the north of the Sermon hall is the Town Square. There are "donations" and "chairs" in the Town Square. To the east of the Town Square is the Armory. In the Armory, there is a "sword", a "shield", "armor" and a "bow". To the north of the Town Square is a nearby road. There is a “knight” there. To the west of the nearby road is a Dungeon. In there is "gold", "jewelry", "gold cups" and a "golden goblet". Deeper in the Dungeon is a secret magician's workshop. In the secret magician's workshop is a "master wizard", "servants", "chickens", "ornate tables" and the "granite kingdom seal". Past the secret magician's workshop is a whipping chamber. In there is "coal" and "the wizard's servant". To the north of the nearby road is a Ruined house. In there are "gold and shiny things". To the east of the Ruined house is a Meadow. The game ends when you go here. All relevant NPCs and Objects are in quotes.''

The output from ChatGPT is as follows,
\begin{itemize}
    \item ``What is the objective of the game in LIGHT?''
\end{itemize}

To provide the goal, we use the prompt ---``Shortly describe what I can do to get gold.''
The output from ChatGPT is as follows,
\begin{itemize}
    \item ``To get gold in the game, you can visit the Sermon hall and find the small sack of gold, visit the wealthy area of town and find the gold bars in the Hillside manor, visit the Dungeon and find gold, jewelry, gold cups, and a golden goblet, and visit the Ruined house and find gold and shiny things. You can also potentially find gold by interacting with the watch maker in the Sermon hall, the knight on the nearby road, and the master wizard in the secret magician's workshop.''
\end{itemize}

To provide the personas, we use the following prompts,
\begin{itemize}
    \item ``I am a brave Adventurer. I am unarmed and carrying nothing. Shortly describe which method suits me best. Explicitly say which items I interact with and where I finish the game.''
    \item ``I am a cowardly and greedy Thief. I am unarmed and carrying nothing. Shortly describe which method suits me best. Explicitly say which items I interact with and where I finish the game.''
    \item ``I am a lazy Bum. I am unarmed and carrying nothing. Shortly describe which method suits me best. Explicitly say which items I interact with and where I finish the game.''
\end{itemize}

The outputs from ChatGPT are as follows,
\begin{itemize}
    \item ``As a brave Adventurer unarmed and carrying nothing, it might be best for you to visit the Armory and obtain a sword, shield, armor, and bow for protection. You can then visit the Dungeon and search for gold, jewelry, gold cups, and a golden goblet to acquire some wealth. After obtaining these items, you can visit the Meadow to finish the game.''
    \item ``As a cowardly and greedy Thief, the best method for you to get gold in the game would likely be to visit the Sermon hall and find the small sack of gold, and then visit the wealthy area of town and find the gold bars in the Hillside manor. You can also potentially find gold by interacting with the watch maker in the Sermon hall, although it is not specified what items you would need to interact with to obtain the gold in this scenario.

To finish the game, you should avoid the Dungeon and the secret magician's workshop, as there may be danger and potential consequences for stealing from these areas. Instead, you should head directly to the Meadow to the east of the Ruined house, as this is where the game ends.''
    \item ``It seems that the most suitable method for acquiring gold for a lazy and unarmed player would be to visit the Ruined house, where there are gold and shiny things that can be easily taken without any special equipment or effort. To reach the Ruined house, the player can start at the Simple Town and head north to the Town Square. From there, they can continue north to the nearby road and then turn west to reach the Ruined house.

Alternatively, the player could try to acquire gold by completing tasks or quests for NPCs in the game. This might require some effort, but it could potentially yield a larger reward. For example, the player could speak to the knight on the nearby road and see if he has any tasks that need to be completed in exchange for gold. However, it is not specified in the description how the player can interact with the NPCs or what tasks they might offer.''
\end{itemize}

Additionally, we also demonstrate ChatGPT is capable of generating the triples from the target KG for LIGHT. Again, we provide context about the environment with the prompt---``I am going to describe a game. Please use this description to answer my next question. In the world of LIGHT, there is a Simple Town. There is a Sermon hall to the east. In the Sermon hall are a "high priest", a "watch maker", a "small sack of gold", a "cross" and "old prayer books". The wealthy area of town is to the north. In the wealthy area of town there is a Hillside manor. In the Hillside manor, there are some "rabbits", a "serving boy", "gold bars", "cloths" and "bottles of liquor". To the north of the Sermon hall is the Town Square. There are "donations" and "chairs" in the Town Square. To the east of the Town Square is the Armory. In the Armory, there is a "sword", a "shield", "armor" and a "bow". To the north of the Town Square is a nearby road. There is a “knight” there. To the west of the nearby road is a Dungeon. In there is "gold", "jewelry", "gold cups" and a "golden goblet". Deeper in the Dungeon is a secret magician's workshop. In the secret magician's workshop is a "master wizard", "servants", "chickens", "ornate tables" and the "granite kingdom seal". Past the secret magician's workshop is a whipping chamber. In there is "coal" and "the wizard's servant". To the north of the nearby road is a Ruined house. In there are "gold and shiny things". To the east of the Ruined house is a Meadow. The game ends when you go here. All relevant NPCs and Objects are in quotes.''

The output from ChatGPT is as follows,

\begin{itemize}
    \item ``The game you described is called "LIGHT" and is set in a town called Simple Town. The town has several locations including a Sermon hall, a wealthy area, a Town Square, an Armory, a nearby road, a Dungeon, a secret magician's workshop, a whipping chamber and a Ruined house. Each location has various NPCs (non-player characters) and objects such as a "high priest", "watch maker", "small sack of gold", "cross", "old prayer books", "rabbits", "serving boy", "gold bars", "cloths", "bottles of liquor", "donations", "chairs", "sword", "shield", "armor", "bow", "gold", "jewelry", "gold cups", "golden goblet", "master wizard", "servants", "chickens", "ornate tables" and the "granite kingdom seal". The game ends when the player reaches the Meadow location.''
\end{itemize}

To seed ChatGPT with knowledge about the world in KG form, we use the prompt---``Convert this game to a knowledge graph of comma-separated triples in the form ("object", "in", "location"), ("npc", "is", "alive"), ("npc2", "is", "dead"), or ("you", "has", "object"). Assume all npcs start "alive". Represent this as ("npc", "is", "alive).''

The output from ChatGPT is as follows,

\begin{itemize}
    \item ''("high priest", "in", "Sermon hall"),
("watch maker", "in", "Sermon hall"),
("small sack of gold", "in", "Sermon hall"),
("cross", "in", "Sermon hall"),
("old prayer books", "in", "Sermon hall"),
("high priest", "is", "alive"),
("watch maker", "is", "alive"),
("rabbits", "in", "Hillside manor"),
("serving boy", "in", "Hillside manor"),
("gold bars", "in", "Hillside manor"),
("cloths", "in", "Hillside manor"),
("bottles of liquor", "in", "Hillside manor"),
("serving boy", "is", "alive"),
("donations", "in", "Town Square"),
("chairs", "in", "Town Square"),
("sword", "in", "Armory"),
("shield", "in", "Armory"),
("armor", "in", "Armory"),
("bow", "in", "Armory"),
("gold", "in", "Dungeon"),
("jewelry", "in", "Dungeon"),
("gold cups", "in", "Dungeon"),
("golden goblet", "in", "Dungeon"),
("master wizard", "in", "secret magician's workshop"),
("servants", "in", "secret magician's workshop"),
("chickens", "in", "secret magician's workshop"),
("ornate tables", "in", "secret magician's workshop"),
("granite kingdom seal", "in", "secret magician's workshop"),
("coal", "in", "whipping chamber"),
("the wizard's servant", "in", "whipping chamber"),
("gold and shiny things", "in", "Ruined house"),
("master wizard", "is", "alive"),
("servants", "is", "alive"),
("the wizard's servant", "is", "alive"),
("knight", "is", "alive")''
\end{itemize}

To prompt ChatGPT to generate the triples in the target KG, we use the following prompts,

\begin{itemize}
    \item ''This is a story that occurs in the same setting. A triple is only changed if the specified object or npc moves to another state. If an action occurs, it is added to the knowledge graph along with its effect. For state, use ("you", relation, object). For action, use ("I", action, object). Use the simplest word you can do describe the action. Say where the player is when the game is finished with ("you", "in", location)

Convert actions to their simplest form. For example, use "get" instead of "grab".

Do not include triples where the action is "go".

Do not include potential actions.

After this story, which triples be added to the knowledge graph? 

``As a brave Adventurer unarmed and carrying nothing, it might be best for you to visit the Armory and obtain a sword, shield, armor, and bow for protection. You can then visit the Dungeon and search for gold, jewelry, gold cups, and a golden goblet to acquire some wealth. After obtaining these items, you can visit the Meadow to finish the game.''''

\item ''This is a story that occurs in the same setting. A triple is only changed if the specified object or npc moves to another state. If an action occurs, it is added to the knowledge graph along with its effect. For state, use ("you", relation, object). For action, use ("I", action, object). Use the simplest word you can do describe the action. Say where the player is when the game is finished with ("you", "in", location)

Convert actions to their simplest form. For example, use "get" instead of "grab".

Do not include triples where the action is "go".

Do not include potential actions.

After this story, which triples be added to the knowledge graph? 

``As a cowardly and greedy Thief, the best method for you to get gold in the game would likely be to visit the Sermon hall and find the small sack of gold, and then visit the wealthy area of town and find the gold bars in the Hillside manor. You can also potentially find gold by interacting with the watch maker in the Sermon hall, although it is not specified what items you would need to interact with to obtain the gold in this scenario.

To finish the game, you should avoid the Dungeon and the secret magician's workshop, as there may be danger and potential consequences for stealing from these areas. Instead, you should head directly to the Meadow to the east of the Ruined house, as this is where the game ends.''''

\item ''This is a story that occurs in the same setting. A triple is only changed if the specified object or npc moves to another state. If an action occurs, it is added to the knowledge graph along with its effect. For state, use ("you", relation, object). For action, use ("I", action, object). Use the simplest word you can do describe the action. Say where the player is when the game is finished with ("you", "in", location)

Convert actions to their simplest form. For example, use "get" instead of "grab".

Do not include triples where the action is "go".

Do not include potential actions.

After this story, which triples be added to the knowledge graph? 

``It seems that the most suitable method for acquiring gold for a lazy and unarmed player would be to visit the Ruined house, where there are gold and shiny things that can be easily taken without any special equipment or effort. To reach the Ruined house, the player can start at the Simple Town and head north to the Town Square. From there, they can continue north to the nearby road and then turn west to reach the Ruined house.

Alternatively, the player could try to acquire gold by completing tasks or quests for NPCs in the game. This might require some effort, but it could potentially yield a larger reward. For example, the player could speak to the knight on the nearby road and see if he has any tasks that need to be completed in exchange for gold. However, it is not specified in the description how the player can interact with the NPCs or what tasks they might offer.''''
\end{itemize}

The outputs from ChatGPT are as follows,

\begin{itemize}
    \item ``("sword", "in", "Armory"),
("shield", "in", "Armory"),
("armor", "in", "Armory"),
("bow", "in", "Armory"),
("gold", "in", "Dungeon"),
("jewelry", "in", "Dungeon"),
("gold cups", "in", "Dungeon"),
("golden goblet", "in", "Dungeon")''
\item ``("small sack of gold", "in", "Semon hall")
("gold bars", "in", "Hillside manor")''
\item ``("gold and shiny things", "in", "Ruined house")''
\end{itemize}

\clearpage
\section{Human Study}
\subsection{Commonsense and Social Norm Knowledge}
\label{app:human_exp1}
\begin{figure}[H]
    \centering
    \includegraphics[width=8cm]{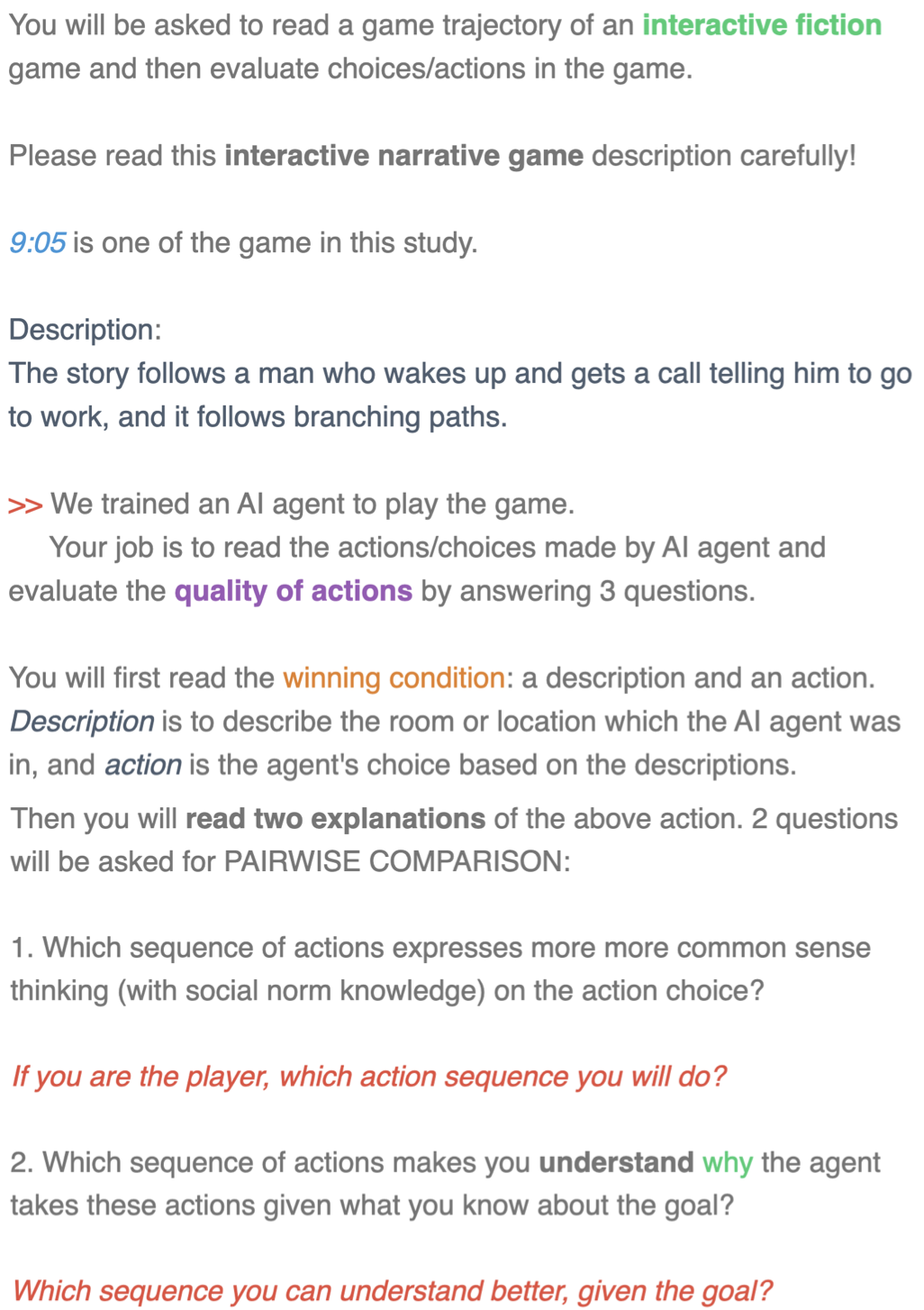}
    \caption{Screenshot of instruction of the human study on evaluating commonsense and social norm knowledge.}
    \label{fig:exp1_instr}
\end{figure}

\begin{figure}[t]
    \centering
    \includegraphics[width=8cm]{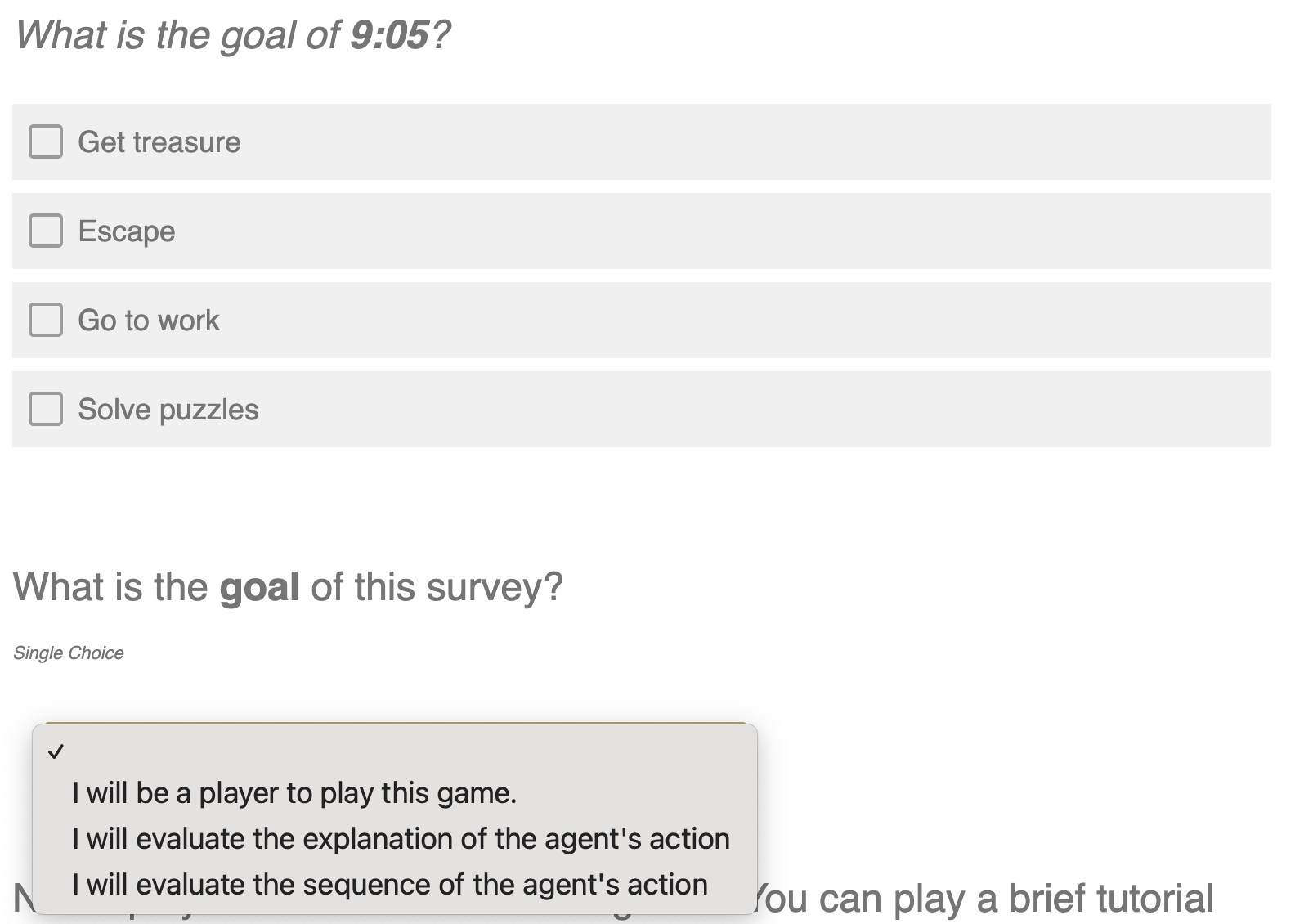}
    \caption{Screen questions of human study -- part 1.}
    \label{fig:exp1_screen1}
\end{figure}

\begin{figure}[t]
    \centering
    \includegraphics[width=8cm]{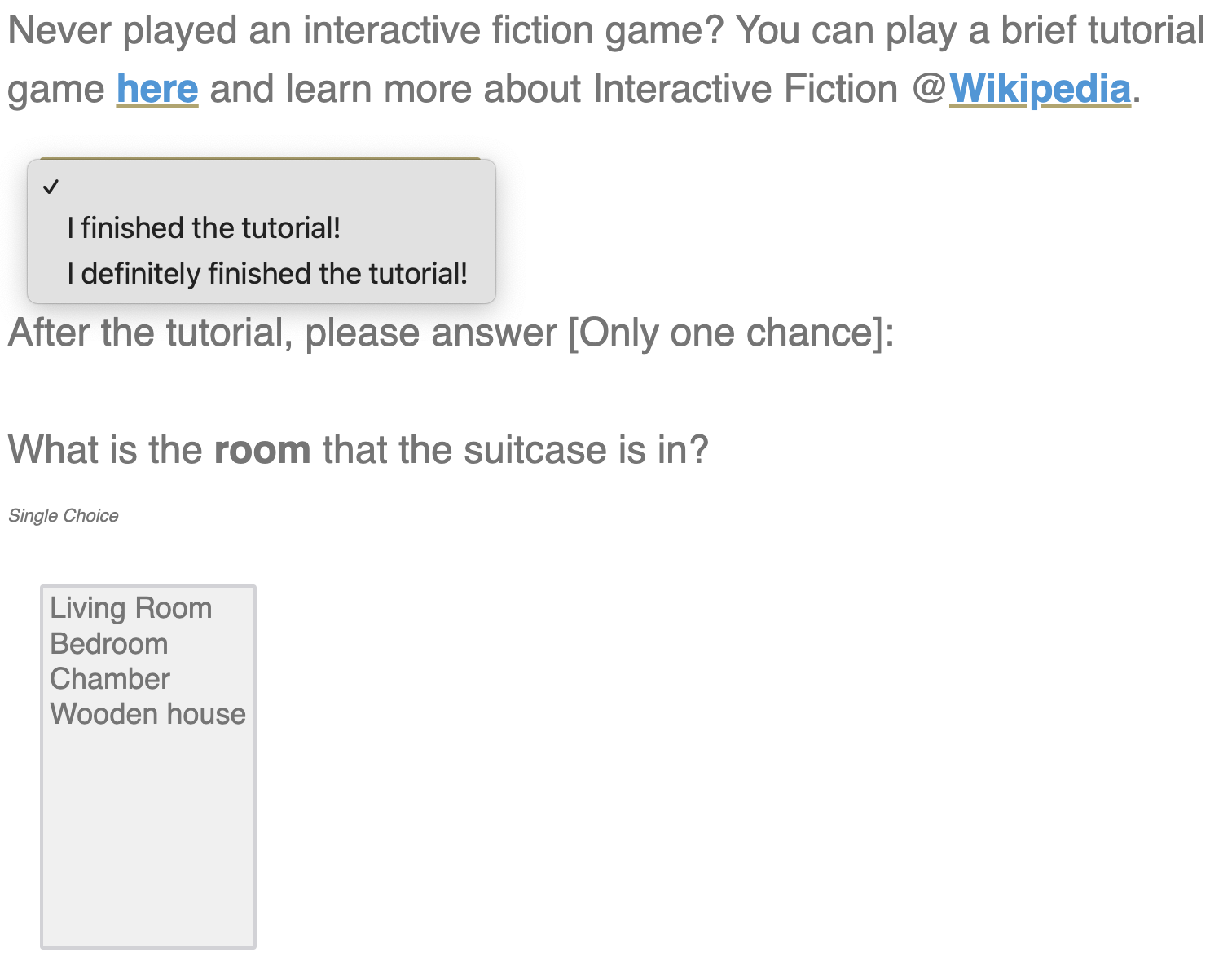}
    \caption{Screen questions of human study -- part 2.}
    \label{fig:exp1_screen2}
\end{figure}

We first show the instruction to them in Figure~\ref{fig:exp1_instr}.
We ask participants a set of screen questions to make sure they understand our task. 
The details can be found in Figure~\ref{fig:exp1_screen1} and \ref{fig:exp1_screen2}.
We conduct our studies using the Cloud Research crowdsourcing platform to interact with Amazon Mechanical Turk \cite{litman2017turkprime}. 
Obtaining at least a bachelor's degree and English as their native language are required to take this study.
Participants are required to pass screening questions and then explain their preferences  of each choice in this study with more than $50$ characters,
which helps filter out low-quality responses and ensures the validity of the study. 
Our study was approved by our Institutional Review Board, and we payed participants the equivalent of $\$15$/hr.

\begin{figure}[H]
    \centering
    \includegraphics[width=8cm]{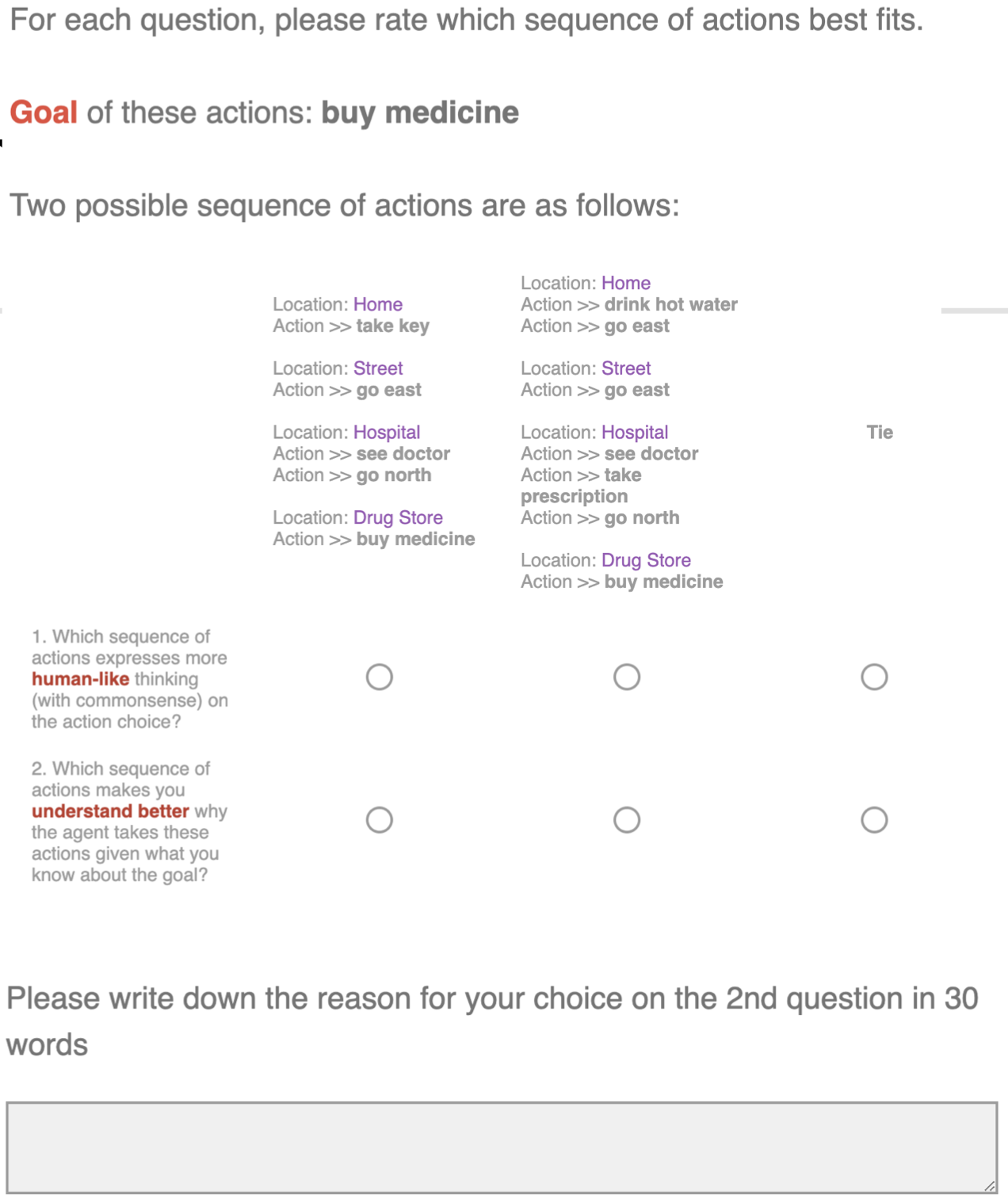}
    \caption{Screenshot of the human study on evaluating commonsense and social norm knowledge.}
    \label{fig:exp_human}
\end{figure}

We assess whether Story Shaping facilitates the expression of commonsense and social norm knowledge in Section~\ref{sec:exp-human}.
We recruited $30$ participants on a crowdsourcing platform.
Each participant reads the winning goal of a randomly chosen game. 
Then they read a pair of game transcript which played by Q*BERT-S and Q*BERT, specifically. Each transcript includes game observations and the corresponding actions.
An example is shown in Figure~\ref{fig:exp_human}.

\subsection{Role Persona Understanding}
\label{app:human_exp2}
\begin{figure}[H]
    \centering
    \includegraphics[width=8cm]{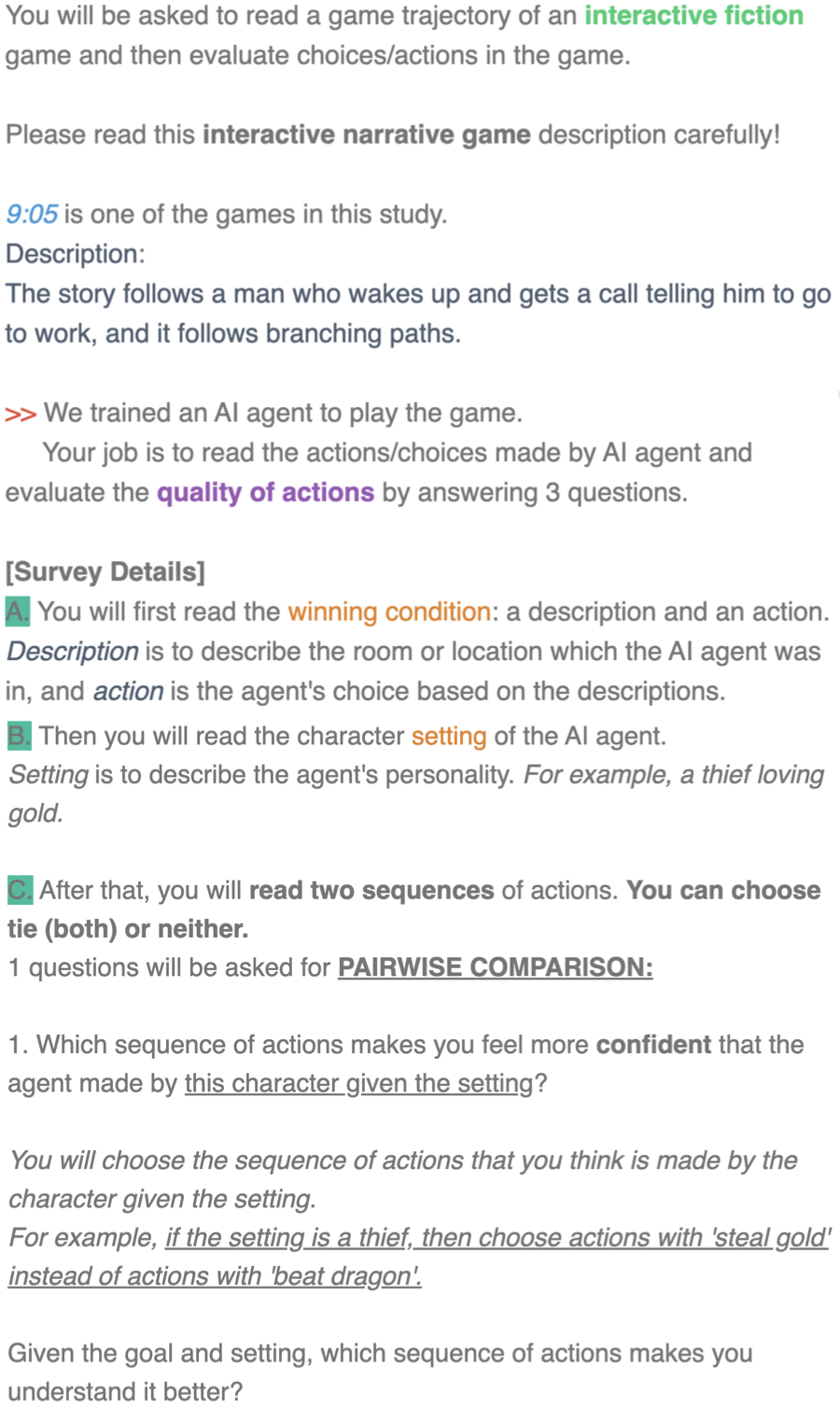}
    \caption{Screenshot of instruction of the human study on evaluating whether Story Shaping has the ability to shape the agent’s behaviors in a way that is identifiable to humans.}
    \label{fig:persona-instr}
\end{figure}

\begin{figure}[H]
    \centering
    \includegraphics[width=8cm]{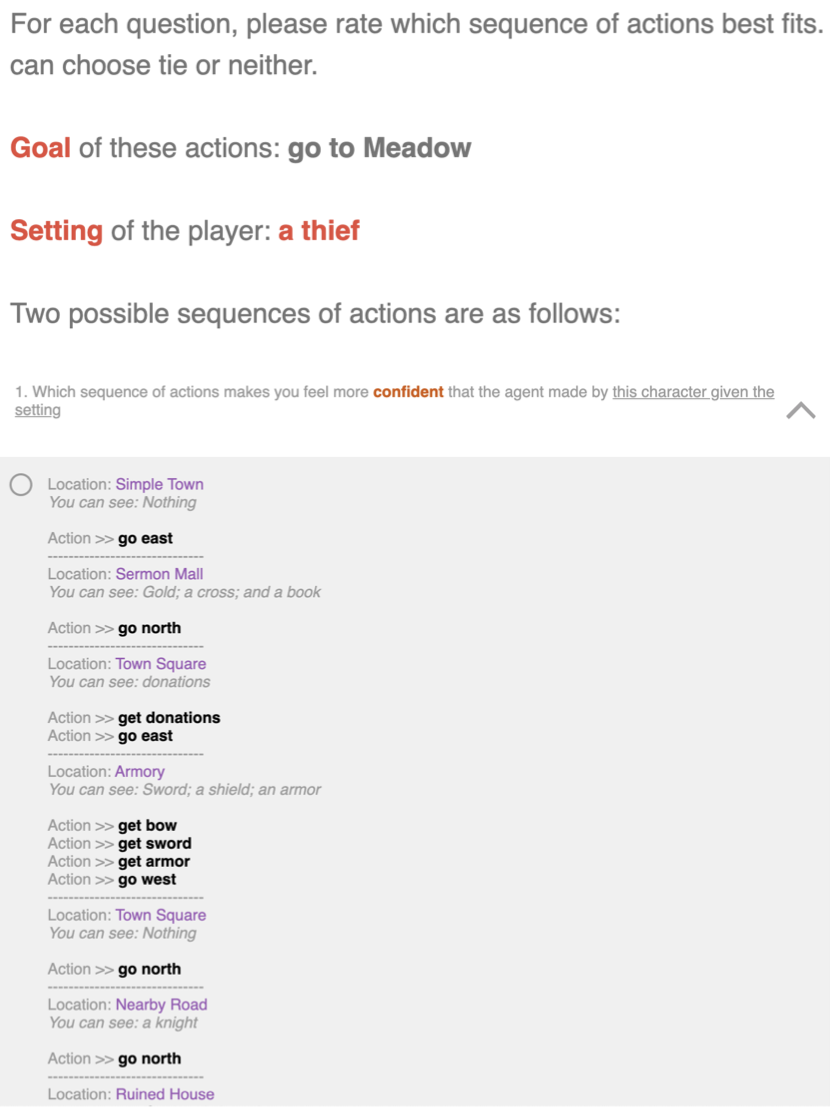}
    \caption{Screenshot of the human study on evaluating whether Story Shaping has the ability to shape the agent’s behaviors in a way that is identifiable to humans --- part 1.}
    \label{fig:exp2-1}
\end{figure}

\begin{figure}[H]
    \centering
    \includegraphics[width=8cm]{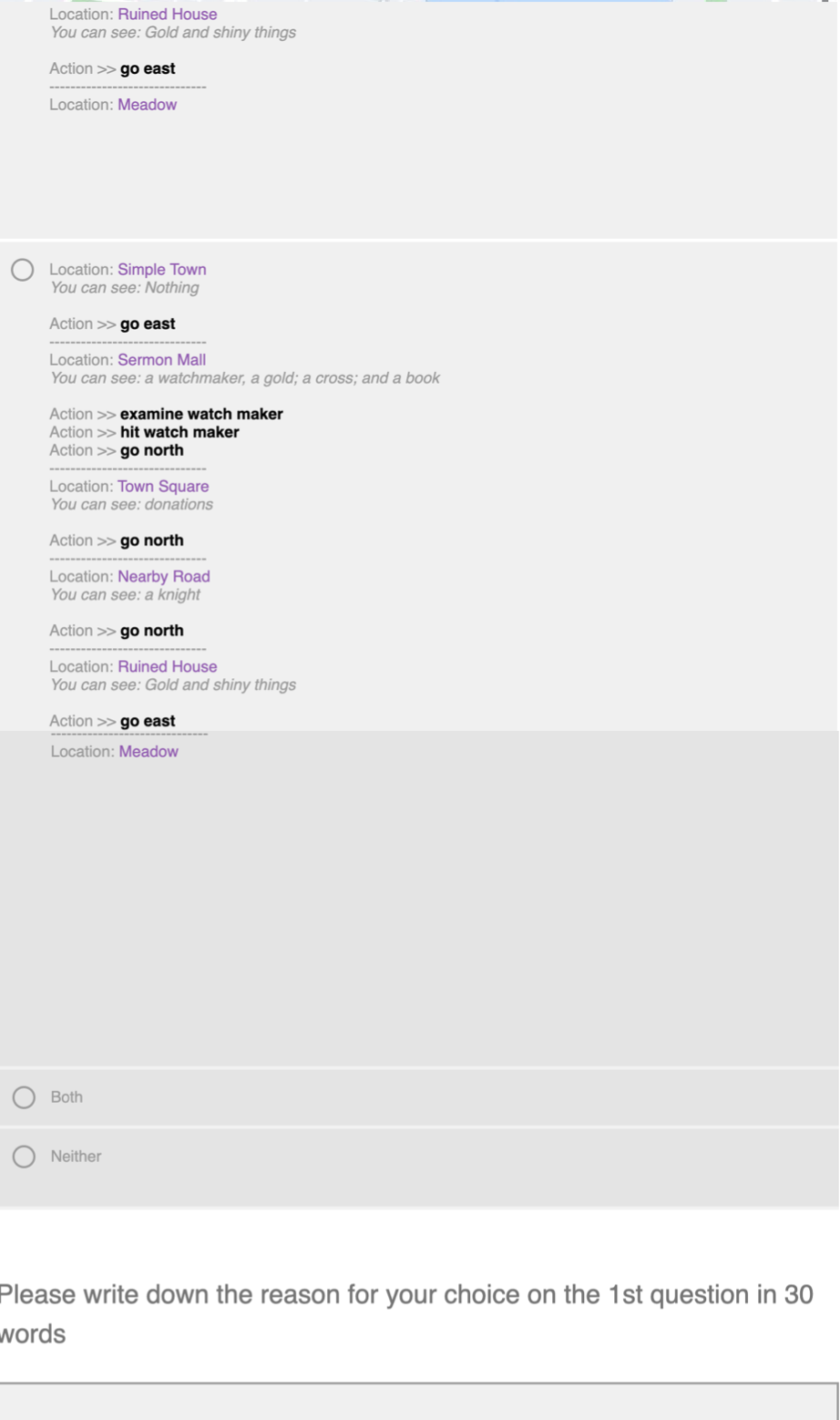}
    \caption{Screenshot of the human study on evaluating whether Story Shaping has the ability to shape the agent’s behaviors in a way that is identifiable to humans --- part 2.}
    \label{fig:exp2-2}
\end{figure}

We first show the instruction to them in Figure~\ref{fig:persona-instr}.
The same screen questions are used here in Figure~\ref{fig:exp1_screen1} and \ref{fig:exp1_screen2}.
We recruited an additional 29 participants. Participants will read a winning goal and be told the agent’s persona (thief, adventurer, thug, bum), then read two game transcripts played by versions of KG-A2C with Story shaping.
The details can be found in Figure~\ref{fig:exp2-1} and \ref{fig:exp2-2}.

\subsection{Robustness}
\label{app:human_exp3}
We replicate the human study in Section~\ref{app:human_exp1}.

\end{document}